\documentclass[10pt]{extarticle}
\usepackage{authblk}

\usepackage[a4paper,bindingoffset=0.2in,%
left=0.8in, right=0.8in, top=0.7in, bottom=0.7in,%
]{geometry}
\usepackage{setspace}
\usepackage{longtable}

\usepackage[pdftex]{color,graphicx}
\usepackage{fullpage}
\usepackage{xspace, wrapfig, multirow}
\usepackage{setspace,amsmath}
\usepackage[draft,ulem=normalem]{changes} 
\usepackage[numbers]{natbib}
\usepackage{url}
\usepackage[normalem]{ulem}

\usepackage{caption}
\usepackage{subcaption}

\usepackage{amsmath}
\usepackage{amssymb}
\usepackage{mathrsfs}
\usepackage{booktabs}
\usepackage[detect-all]{siunitx}
\usepackage{etoolbox}
\newrobustcmd*{\bftabnum}{%
	\bfseries
	\sisetup{output-decimal-marker={\textmd{.}}}%
}

\usepackage{xcolor,mdframed}
\usepackage{array}
\usepackage{float}
\usepackage{color,soul}

\usepackage{pdflscape}
\usepackage{afterpage}
\usepackage{capt-of}
\usepackage{lipsum}
\usepackage{rotating}
\usepackage{threeparttable}
\usepackage{etoolbox}
\appto\TPTnoteSettings{\footnotesize}
\usepackage{adjustbox}
\usepackage{pbox}

\usepackage{verbatim}

\usepackage{algorithm}
\usepackage{algcompatible}

\usepackage{tabularx}

\usepackage{makecell}
\usepackage{cellspace} %
\algnewcommand\INPUT{\item[\textbf{Input:}]}%
\algnewcommand\OUTPUT{\item[\textbf{Output:}]}%
\algnewcommand\BREAK{\State \textbf{break}}%
\algnewcommand\BULLET{\item[$\bullet$]}%

\usepackage{amsthm}
\newtheorem{cor}{Corollary}

\newtheorem{theorem}{Theorem}
\newtheorem{assume}{Assumption}
\newtheorem{definition}{Definition}
\newtheorem{lemma}{Lemma}

\newcolumntype{P}[1]{>{\centering\arraybackslash}p{#1}}

\allowdisplaybreaks

\title{A Mathematical Programming Approach for Integrated Multiple Linear Regression Subset Selection and Validation}

\author[1]{Seokhyun Chung\footnote{seokhc@umich.edu}}
\affil{Industrial \& Operations Engineering, University of Michigan, Ann Arbor, MI, USA}
\author[2]{Young Woong Park\footnote{ywpark@iastate.edu; corresponding author}}
\affil{Ivy College of Business, Iowa State University, Ames, IA, USA}
\author[3]{Taesu Cheong\footnote{tcheong@korea.ac.kr}}
\affil{School of Industrial Management Engineering, Korea University, Seoul, South Korea}

\begin{document}
\renewcommand{\tabcolsep}{4pt}

\singlespacing

 \maketitle

\begin{abstract}
Subset selection for multiple linear regression aims to construct a regression model that minimizes errors by selecting a small number of explanatory variables. Once a model is built, various statistical tests and diagnostics are conducted to validate the model and to determine whether the regression assumptions are met. Most traditional approaches require human decisions at this step. For example, the user adding or removing a variable until a satisfactory model is obtained. However, this trial-and-error strategy cannot guarantee that a subset that minimizes the errors while satisfying all regression assumptions will be found. In this paper, we propose a fully automated model building procedure for multiple linear regression subset selection that integrates model building and validation based on mathematical programming. The proposed model minimizes mean squared errors while ensuring that the majority of the important regression assumptions are met. We also propose an efficient constraint to approximate the constraint for the coefficient $t$-test. When no subset satisfies all of the considered regression assumptions, our model provides an alternative subset that satisfies most of these assumptions. Computational results show that our model yields better solutions (i.e., satisfying more regression assumptions) compared to the state-of-the-art benchmark models while maintaining similar explanatory power.

\end{abstract}

\section{Introduction}
\label{S:1}
Regression analysis is one of the most popular forms of statistical modeling for analyzing the relationship between variables. Due to its interpretability and simplicity, \emph{multiple (multivariate) linear regression} has been the most commonly used model for several decades, with a variety of fields employing it to handle prediction tasks. Multiple linear regression seeks to identify the most suitable linear relationship for several explanatory variables and a response variable. A multiple linear regression model for $m$ explanatory variables can be represented by the linear equation 
\begin{equation*}
\underset{n \times 1}{\boldsymbol{b}} =   \underset{n \times m}{\boldsymbol{A}}\>\underset{m \times 1}{\boldsymbol{x}} + \boldsymbol{e},
\end{equation*} where $\boldsymbol{A}$ represents the data matrix corresponding to $m$ explanatory variables with $n$ observations, $\boldsymbol{b}$ denotes the data matrix corresponding to the values of the response variable, $\boldsymbol{x}$ is the matrix of estimated parameters corresponding to the coefficients of the variables, and $\boldsymbol{e}$ refers to the errors between the predictions of the linear model and the response variable. We further assume that data matrices $\boldsymbol{A}$ and $\boldsymbol{b}$ are already standardized with their respective means and standard deviations, and thus, without loss of generality, we can remove the intercept from the model above. It then makes sense that $\boldsymbol{x}$ should be chosen to minimize the error. 

Subset selection is a procedure in which a subset of $m$ explanatory variables is selected to reduce model complexity while maintaining explanatory power. \citet{Guyon03} and \citet{james2013introduction} demonstrate that a reduced subset can provide faster, more cost-effective predictors, along with a better understanding of the underlying process that generated the data. A reduced subset can also prevent over-fitting in the presence of too many irrelevant or redundant features \citep{yu2004efficient} and can help users to more readily understand the results \citep{karegowda2010feature}.

Many studies have been conducted on subset selection in the literature. Stepwise selection, including forward selection, backward elimination, and their combination, are the most popular algorithms thanks to their simple implementation and fast computing time. However, the solutions found by these algorithms are often low quality due to their greedy characteristics. Therefore, in order to improve subset selection, more complicated algorithms have been proposed. Meta-heuristic algorithms are presented in \citet{zhang2002optimal} (Tabu Search), \citet{siedlecki1989note} (genetic algorithm), and \citet{hafiz2018two} (particle swarm). Subset selection procedures based on statistical or machine learning methods have also been carried out. For example, a Bayesian approach to subset selection is taken by \citet{mitchell1988bayesian}, while \citet{george1993variable} suggest a selection procedure based on the superiority of each subset estimated from posterior probabilities given by Gibbs sampling. \citet{genuer2010variable} point out that random forests can be employed in a strategy that involves the ranking of explanatory variables, which provides insight for the selection of the variable(s). In addition, other research on machine learning algorithms for subset selection has been conducted by \citet{castellano2000variable} (neural network), \citet{rakotomamonjy2003variable} (support vector machine), and \citet{zheng2018feature} (information entropy-based objective function). By assigning an $L_1$ penalty to coefficients, the least absolute shrinkage and selection operator (LASSO) is introduced by \citet{tibshirani1996regression}. Although there have been a few decades since LASSO was first devised, it is still one of the most commonly used regression subset selection methods. Note that most of the methods above are heuristic, which means they do not necessarily give the subset with the minimum mean squared error (MSE).

Once a multiple linear regression model with a selected subset is established, the built model is validated through various statistical tests \cite{neter1996applied}. This validation is essential when the model is used as an explanatory model, which is the usual case for linear regression. For example, if a coefficient in multiple linear regression fails to pass the statistical significance test, the interpretation of data using the constructed model is not  statistically justified. If the residual plots show violation of the regression assumptions, the validity of the model is hurt. It is thus crucial to building a model that achieves a small MSE as well as statistical validity.

\subsection{Related work}
 Although the best subset selection problem can be formulated as mixed integer quadratic programming (MIQP) when ordinary least squares (OLS) is used, the statistical community has not paid it close attention because it requires significant computational time for practical implementation. However, mathematical programming approaches to find the best subset minimizing MSE have gained increased attention in the literature recently. \citet{bertsimas2016OR} have pointed out that the computing power of mixed integer programming (MIP) solvers has increased at a rapid rate in the last 25 years, while also emphasizing the significant progress in the development of exact algorithms to solve integer programs. These recent developments have made integer programming a key method for finding the best subset. \citet{konno2009choosing} introduce an MIQP formulation to find the best subset with a fixed number of variables for multiple linear regression with the minimum sum of squared error (SSE). \citet{konno2010multi} propose a multistep method to generate a nearly optimal solution in a shorter time. Based on the formulation of \citet{konno2009choosing}, \citet{bertsimas2016best} suggest obtaining tight big-$M$ values in the indicator constraint to improve computing speed. They also introduce an algorithm to achieve an initial solution for a warm-start, effectively reducing the computing time. The choice of a best subset for logistic regression via MIP using piecewise linear approximation is also presented in \citet{sato2016feature}. Models and algorithms that do not require the number of selected variables to be fixed have also been proposed. \citet{miyashiro2015mixed} introduce mixed integer second-order cone programming formulations that do not restrict the cardinality of the subsets. \citet{miyashiro2015subset} set Mallows' $C_p$ as the goodness of fit for the model and formulate the subset selection as MIP. In their work, Mallows' $C_p$ includes the number of selected variables and hence the cardinality of the subset need not be fixed when Mallows' $C_p$ is used as a measure of the goodness of fit. \citet{park2013subset} consider an MIQP formulation for subset selection when the shape of the data is not only ordinary (i.e., $m<n$) but also for high dimensional (i.e., $m>n$) variable selection when the number of selected variables is not fixed. They also provide a proof for big-$M$ being an upper bound on coefficients and an efficient heuristic algorithm for cases when $m<n$. Note that these studies only consider several goodness-of-fit measures to obtain an optimal subset. \citet{gomez18} recently developed a mixed-integer fractional programming model for criteria such as the Akaike Information Criterion (AIC), the Bayesian Information Criterion (BIC), and the Hannan-Quinn Information Criterion (HQIC). They further strengthen their formulations by exploiting the normal equations underlying the optimization problem in computing.

While numerous studies have focused on regression subset selection, few have considered regression assumptions and diagnostics in their solution approach. Diagnostics are indeed an essential component in building a good explanatory regression model to explain the linear relationship between the response and explanatory variables. While regression models can be used to make predictions and be evaluated based on their prediction performance, in this study, we focus on improving the interpretability and explanatory power of a regression model through regression diagnostics. \citet{bertsimas2016OR} suggest a bootstrap-based algorithmic approach that iteratively solves MIP problems to obtain a desirable model. They use penalized objective functions and constraints for sparsity, multicollinearity reduction, robustness to outliers, and statistical significance. \citet{tamura2016best} propose MIQP for best subset selection while eliminating multicollinearity from linear regression models. \citet{carrizosaaenhancing} study a mathematical model for subset selection in order to construct a linear regression model with the significant coefficients and small multicollinearity by constraints performing shrinkage of the coefficients. \citet{kim2018collinear} consider group-wise multicollinearity in best subset selection framework and developed a modified discrete first-order algorithm to reduce multicollinearities in the selected subset. Recently, \citet{Bertsimas2019Accounting} proposed an MIP-based framework based on \citet{bertsimas2016OR} and accounted for significance and multicollinearity. They also propose an asymptotic normality result, which works independent of the regression normality assumption.

Our model differs from \citet{bertsimas2016OR} and \citet{carrizosaaenhancing} in that our algorithm finds a solution that satisfies the tests and diagnostics with only one call to an MIP solver, whereas they use an iterative method that calls an MIP solver multiple times or heuristic observations for modeling. When the regression normality assumption is met, our approach for significance test is equivalent to the model of \citet{Bertsimas2019Accounting}; our work has been conducted independently and simultaneously with \citet{Bertsimas2019Accounting}.

In this study, we propose a fully automated model building and validation procedure for multiple linear regression via a mathematical programming-based algorithm that minimizes MSE while guaranteeing both (i) the statistical significance of the regression coefficients and  (ii) regression assumptions and diagnostics. The proposed model can replace the traditional iterative validation and diagnostics steps, and, with the help of lazy constraint technique, successfully returns the best subset with significant coefficients and a regression assumption-satisfying model in most cases. In addition, for the statistical significance test that is difficult to formulate as an easier linear constraint, we propose an explicit constraint to approximate the exact constraint for the coefficient $t$-test. The explicit constraint is formulated as a linear constraint and directly added to the model upfront to  further reduce the solution time and size of the branch-and-bound tree explored. Our model is also capable of providing an alternative solution when there is no subset that satisfies all of the considered regression assumptions and tests. 

\subsection{Contribution}

The contributions of our paper can be summarized as follows:

\begin{itemize}
\item We provide a mathematical programming-based algorithm that allows a regression model to be built that satisfies the majority of statistical tests and diagnostics. Note that it is not trivial to incorporate this model validation step into other popular subset selection approaches such as LASSO and stepwise selection. To the best of our knowledge, residual-based diagnostics are incorporated into the model for the first time in the literature. The experimental results show that our model identifies a subset that satisfies all of the considered tests and assumptions within a reasonable time while minimizing adjusted $R$ squared.

While an MIP solver is used to solve the problem, our algorithm is unique in the sense that it does not require multiple iterative calls to the solver. The experimental results demonstrate that our algorithm significantly speeds up the solution search in comparison to existing iterative methods, resulting in the success of finding better solutions.

\item We propose an efficient explicit constraint that significantly reduces the solution space. The constraint is a relaxed version of the constraint we propose for the regression coefficient $t$-test. This relaxed constraint is added to the model before the branch-and-bound algorithm to exclude the solutions (subsets) that significantly violate the $t$-test from the branch-and-bound search tree. The computational experiment shows that the relaxed constraint reduces the solution time and size of the search tree.

\item  We present a logical procedure to find a near-feasible solution when no subset satisfies the tests and diagnostics or our model fails to find a feasible solution within the given time limit. To the best of our knowledge, our work is the first attempt that provides good alternative solutions in the case, which indeed often arises in practice. The proposed procedure mimics the typical steps used to build a linear regression model. The experimental results show that our procedure produces higher quality subsets than do the benchmarks.
\end{itemize}

\subsection{Structure of the paper}

The paper is structured as follows. In Section \ref{S:2}, we discuss several characteristics of a desirable multivariate linear regression model that incorporates transformed variables and meets essential regression assumptions. In Section \ref{S:3}, a mathematical programming approach for the best subset which reflects the features discussed in Section \ref{S:2} is presented. A logical procedure for obtaining an alternative solution when the model is infeasible is proposed in Section \ref{S:4}. The results of computational experiments are presented in Section \ref{S:5}, followed by conclusions in Section \ref{S:6}.

\section{Model Validation for Multiple Linear Regression}\label{S:2}

\begin{figure}[h!]\centering
	\includegraphics[page=1, width=\textwidth]{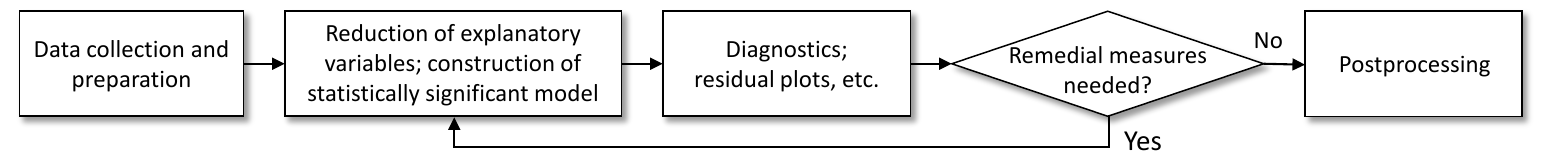}
	\caption{Strategy for linear regression model construction \citep{neter1996applied}}\label{fig:0}
\end{figure}

Subset selection in multiple linear regression aims to find a subset of explanatory variables that gives small fitting errors. At the same time, the model needs to satisfy the assumptions of linear regression from a statistical perspective. To find  such subset, \citet{neter1996applied} suggests building a regression model by repetitively checking the assumptions and diagnostics. Figure \ref{fig:0} summarizes this strategy. After preprocessing the collected data, several explanatory variables are selected to build a statistically significant model. Diagnostics are then conducted on the model. If the model satisfies all assumptions, it is forwarded to the postprocessing step. Otherwise, another subset of explanatory variables is selected and tested. As a result, constructing a useful regression model requires many diagnostics tests. In this section, we discuss three popular techniques and tests that will be included in the mathematical formulation proposed in Section \ref{S:3}.

 \subsection{Transformation of explanatory variables}\label{S:2_1}
 When the explanatory and response variables have a non-linear relationship, the non-linear trend can be observed in the residual plot. To remedy this issue, $\log$ or other nonlinear transformation can be used for the explanatory variables. When the transformed variables are added to the set of explanatory variables, at most one between the original and transformed explanatory variables can be included in the subset simultaneously to avoid multicollinearity. If the transformed explanatory variables work properly, the non-linear relationship issue in the residual plot can be fixed. Therefore, we generate transformed data as a fixed candidate set of selected variables. The MIQP model presented in Section \ref{S:3_2} has a constraint that selects at most one of the original and transformed variables from the pair.

 \subsection{Statistical tests for linear regression parameters}\label{S:2_2}
 
One of the popular tests in the model evaluation step is the statistical significance test for regression coefficients. This test checks if the estimated regression coefficients are non-zero using Student's $t$-distribution. Let $s(\boldsymbol{A})_j$ and $\hat{x}_j$ be the standard deviation and estimated coefficients, respectively, of explanatory variable $j \in \{1,...,k\}$. Then, the coefficients follow Student's $t$-distribution.
 
 \begin{equation}
 \label{eq:ttest_1}
 \frac{\hat{x}_j - x_j}{s(\boldsymbol{A})_j} \sim t_{1-\frac{\alpha}{2} ,n-k-1}, \qquad j={1,...,k}.
 \end{equation}
 
 \noindent Our interest is to test whether coefficient $\hat{x}_j$ is equal to zero. We conclude that linear relationship between the response variable and the $j$-th explanatory variable is not statistically significant if $\hat{x}_j$ is close to zero with a large standard deviation. On the other hand, if $\hat{x}_j$ is large enough, we conclude that there is a significant linear relationship. Formally, the following hypothesis can be tested: $H_0: x_j = 0,  H_1: x_j \neq 0.$ Thus, to reject the null hypothesis, the following inequality must hold:
 \begin{equation}
 \label{eq:ttest_2}
 \left| \frac{\hat{x}_j}{s(\boldsymbol{A})_j} \right| \geq t_{1-\frac{\alpha}{2} ,n-k-1}, \qquad j={1,...,k}.
 \end{equation}
 
 We will discuss how this requirement can be handled with a mathematical programming approach in Section \ref{S:3_3} and Section \ref{S:3_4}.

 \subsection{Model validation with residual plots}
 \label{S:2_3}
One of the key assumptions of linear regression is that errors have constant variance over fitted values or over explanatory variables. This can be verified by drawing residual plots. The residuals have constant variance if they form a band-like region with constant width over the entire range of fitted values. However, linear models often do not satisfy the assumption. The violation of residual assumptions can induce the biased estimation of standard errors that may result in invalid inferences \cite{breusch1979simple}. Thus, once a linear regression model is constructed, it is critical to check the diagnostic plots of the residuals.

 \begin{figure}[h!]
 \centering
 \begin{subfigure}[b]{0.24\textwidth}
     \includegraphics[page=1, width=\textwidth]{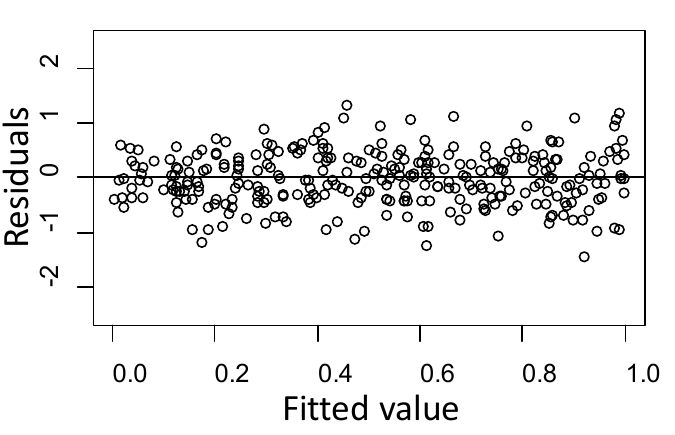}
     \caption{Ideal case}\label{fig:1a}
 \end{subfigure}
 \hfill
 \begin{subfigure}[b]{0.24\textwidth}
     \includegraphics[page=2, width=\textwidth]{resid-example.pdf}
     \caption{Correlated}\label{fig:1b}
 \end{subfigure}
 \hfill
 \begin{subfigure}[b]{0.24\textwidth}
     \includegraphics[page=3, width=\textwidth]{resid-example.pdf}
     \caption{Heteroscedasticity}\label{fig:1c}
 \end{subfigure}
 \hfill
 \begin{subfigure}[b]{0.24\textwidth}
     \includegraphics[page=4, width=\textwidth]{resid-example.pdf}
     \caption{Nonlinearity}\label{fig:1d}
 \end{subfigure}
 \caption{Examples of residual plot}\label{fig:1}
 \end{figure}

 Figure \ref{fig:1} shows representative plots of residuals versus fitted values. In Figure \ref{fig:1a}, the variance of the residuals seems to be constant over the range of fitted values and thus the generated model satisfies the key regression assumption. However, this is not the case for the other examples in Figure \ref{fig:1}. Figure \ref{fig:1b} displays a positive \textit{correlation} between the residuals and fitted values, Figure \ref{fig:1c} shows \emph{heteroscedasticity}, in which the variance of the residuals increases as the fitted values increase, and Figure \ref{fig:1d} presents a \textit{non-linear} relationship between the residuals and fitted values. The plot in Figure \ref{fig:1a} is ideal, whereas the latter three cases in Figure \ref{fig:1} are problematic, need to be fixed, and frequently observed when implementing linear regression models for real-world problems. We next describe how to detect these cases without the visual aids.
 
 \noindent{\textit{\textbf{Correlation in residuals}}} To detect the situation depicted in Figure \ref{fig:1b}, a simple linear regression model, called an \emph{auxiliary model}, is constructed where the explanatory and response variables are the fitted values and residuals in the plot, respectively. If the estimated slope is close to zero, then we conclude the model does not suffer from linearity. If the estimated slope is far from zero, then we conclude the model violates the residual assumption. A simple hypothesis test can be used to check whether the estimated slope is zero. To describe this test mathematically, let $\hat{\boldsymbol{b}}$ and $\hat{\boldsymbol{e}}$ be the vector of fitted values and residuals, respectively. Then, we establish an auxiliary model 
$
\hat{\boldsymbol{e}} = x^\text{aux}_1 \boldsymbol{1} + x^\text{aux}_2 \hat{\boldsymbol{b}}  + \boldsymbol \epsilon    $
where $x^\text{aux}_1$ and $x^\text{aux}_2$ represent the estimated intercept and coefficient, respectively, and $\boldsymbol \epsilon$ is the error of the model. To check whether $x^\text{aux}_2$ is statistically significant, the following hypothesis test is conducted: $H_0: x^\text{aux}_2 = 0$ and $H_1: x^\text{aux}_2 \neq 0$. It is then followed by a usual $t$-test. Note that a failure to reject $H_0$ indicates the linearity assumption is met.
 
\begin{figure}[h!]
  \centering
 \begin{subfigure}[b]{0.3\textwidth}
     \includegraphics[page=1, width=\textwidth]{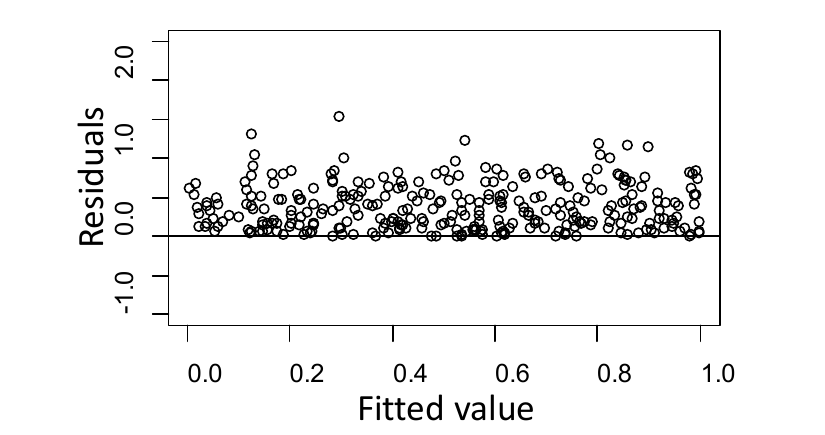}
     \caption{Ideal case}\label{fig:2a}
 \end{subfigure}
 \begin{subfigure}[b]{0.3\textwidth}
     \includegraphics[page=2, width=\textwidth]{resid-abs-exp.pdf}
     \caption{Heteroscedasticity}\label{fig:2b}
 \end{subfigure}
 \caption{Examples of absolute residual plot}\label{fig:2}
 \end{figure}

 \noindent{\textit{\textbf{Heteroscedasticity}}} To detect heteroscedasticity (Figure \ref{fig:1c}), two statistical tests need to be performed. The first test, referred to as an \emph{absolute residual test} in this paper, is identical to the previous test except for the fact that the values of the response variable in the auxiliary model are the absolute values of the residuals. Observe that if the negative side of Figure \ref{fig:1c} is flipped to overlap the positive side, an increasing trend in the residuals is observed as illustrated in Figure \ref{fig:2b}. On the other hand, the same procedure for the ideal case of Figure \ref{fig:1a} creates the constant variance of the absolute residuals over fitted values in Figure \ref{fig:2a}. Thus, if a linear trend is detected from the auxiliary model of the absolute residuals, we can assume that the residuals of the linear model have heteroscedasticity.  Another widely used diagnostic tool for heteroscedasticity is the \textit{Breusch-Pagan test} \citep{breusch1979simple}. To prevent being overly rigorous, we conclude that the heteroscedasticity exists and the residual assumption is violated only if both proposed tests indicate heteroscedasticity. In Section \ref{S:3_4}, our mathematical programming-based diagnostic approach is presented to capture the features of heteroscedasticity. 
 
 \noindent{\textit{\textbf{Nonlinearity in residuals}}} To remedy the nonlinear trend in the residual plot in Figure \ref{fig:1d}, we consider both original and transformed explanatory variables, as explained in Section \ref{S:2_1}, while we restrict the model to select at most one among the original and transformed variables.

 We note here that other tests such as the $F$-test for the model and the multicollinearity test for the pairwise coefficients can be incorporated into our framework proposed in Section \ref{S:3}.

 \section{Integrated Model Building and Validation Procedure via Mathematical Programming}\label{S:3}

In this section, we develop an automated procedure based on mathematical programming models that minimizes SSE to select a subset of a fixed number of explanatory variables and include statistical tests and diagnostics for multivariate linear regression. In Section \ref{S:3_1}, we introduce the base model from the literature, which does not consider statistical test or diagnostics. In Section \ref{S:3_2}, we extend the base model to include log-transformed explanatory variables. In Section \ref{S:3_3}, we propose constraints to remove statistically insignificant coefficients. In Section \ref{S:3_4}, we present the final algorithm, which considers all of the remaining tests and diagnostics. 

 \subsection{Base model}\label{S:3_1}
 We start with a basic model similar to that proposed by \citet{Hoyt78} and all data matrices are standardized. The sets, parameters, and decision variables used are as follows. 
 
 \vspace{0.3cm}
 
 \noindent\textbf{Sets and Parameters}
 
 \begin{longtable}{p{0.8cm} r p{15cm}}
 $(n, m)$                 & : & number of observations and explanatory variables \\
 $(\mathcal{I}, \mathcal{J})$       & : & index set of observations, $\mathcal{I}=\{1,...,n\}$, and explanatory variables, $\mathcal{J}=\{1,...,m\}$ \\
 $k$                 & : & number of selected explanatory variables\\
 $\boldsymbol{A}$    & : & standardized data matrix corresponding to explanatory variables,  \mbox{$\boldsymbol{A} = [\boldsymbol{A}_j] = \lbrack a_{ij} \rbrack \in \mathbb{R}^{n\times m}$ } \\
 $\boldsymbol{b}$        & : & standardized data matrix corresponding to response variables,  $\boldsymbol{b} = \lbrack b_{i} \rbrack \in \mathbb{R}^{n} $
 \end{longtable}
 \noindent\textbf{Decision Variables}\vspace{0.3cm}\\
  \begin{tabular}{p{0.1cm} r p{13.7cm}}
 $x_j$       & :& coefficient of the $j$-th explanatory variable, $\forall j \in \mathcal{J}$\\
 $e_i$       & :& error (i.e., residual) between the $i$-th observation and its prediction value, $\forall i \in \mathcal{I}$ \\
 $z_j$       & :& $1$ if the $j$-th explanatory variable is selected; $0$ otherwise, $\forall j \in \mathcal{J}$
 \end{tabular}
\vspace{0.3cm}
 
 \addtocounter{table}{-1}
 
 Using the parameters and decision variables above, the following basic mathematical programming model can be formulated.
 \begin{subequations}
 \begin{align}
 \quad\quad  \text{minimize }		& \quad \sum_{i\in \mathcal{I}} e_i^2 \label{f:1_0} \\
       \text{subject to }	& \quad e_i = \sum_{j\in \mathcal{J}} a_{ij}x_{j} - b_{i}, \qquad	&& \forall i \in \mathcal{I}, \label{f:1_1} \\
       						& \quad -M z_{j} \leq x_{j} \leq M z_{j}, 			 	\qquad	&& \forall j \in \mathcal{J}, \label{f:1_2} \\
                             & \quad \sum_{j\in J} z_{j} = k,					 	\qquad	&& \label{f:1_3} \\
                             & \quad x_{j} \text{ unrestricted}, \quad z_{j} \in \{ 0, 1 \},	 				\qquad && \forall j \in \mathcal{J},\label{f:1_4} \\
                             & \quad e_{i} \text{ unrestricted},						\qquad && \forall i \in \mathcal{I}.\label{f:1_6}
 \end{align}\label{f:1}
 \end{subequations}
   The basic model (\ref{f:1}) minimizes the SSE of a multivariate linear regression model with a fixed $k$. Note that although $\text{MSE} =  \frac{\sum_{i\in \mathcal{I}} e_i^2}{n-k-1}$ is not directly minimized in \eqref{f:1_0},  minimizing SSE $=\sum_{i\in \mathcal{I}} e_i^2$ in \eqref{f:1_0} is equivalent to minimizing MSE because $k$ is fixed as a given parameter. Constraint \eqref{f:1_1} defines the residuals, and constraint \eqref{f:1_2} indicates that if an explanatory variable is not selected, then the coefficient of the variable must be zero. Lastly, constraint \eqref{f:1_3} ensures that the number of selected variables is $k$.

 The mathematical model (\ref{f:1}) can be converted into an MIQP using a popular linearization technique. Instead of unrestricted continuous decision variables $x_j$ and $e_j$, non-negative variables $x_j^+$, $x_j^-$, $e_j^+$, and $e_j^-$ are used, where $e_i = e_i ^+ - e_i^-$ and $x_i = x_i ^+ - x_i^-$. By plugging these in, the following MIQP can be obtained:
 
 \allowdisplaybreaks
\begin{align}
 \quad\quad \text{minimize }	& \quad \sum_{i\in \mathcal{I}}{ \left( {e_{i}^{+}}^2 + {e_{i}^{-}}^2 \right)} \notag\\
     \text{subject to }	& \quad e_i^+ - e_i^- = \sum_{j\in \mathcal{J}} a_{ij}\left( x_{j}^+ - x_{j}^- \right) - b_{i},
						 	\>	&& \forall i \in \mathcal{I}, \notag\\
     							& \quad x_{j}^+ + x_{j}^- \leq M z_{j}, 			\>	&& \forall j \in \mathcal{J}, \label{f:2}\\
                            	& \quad \sum_{j\in J} z_{j} = k,					\>	&&  \notag\\
                             	& \quad x_{j}^+, x_{j}^- \geq 0, \quad z_{j} \in \{ 0, 1 \},  					\> &&  \forall j \in \mathcal{J}, \notag\\
                             	& \quad e_{i}^+, e_{i}^- \geq 0,  					\> && \forall i \in \mathcal{I}.\notag
  \end{align}

 Recall that the main purpose of our study is to build a multivariate linear regression model that considers diagnostics. To meet this goal, we now extend the base mathematical model (\ref{f:2}) to include the important diagnostic tests presented in Section \ref{S:2} as constraints.

 \subsection{Inclusion of log-transformed explanatory variables}\label{S:3_2}
 We first discuss how to include log-transformed explanatory variables in the model. Because an explanatory variable and its log-transformation are highly correlated, we need to prevent both variables from being selected simultaneously, as discussed in Section \ref{S:2_1}. Let us then define new parameters and sets.
 
 \noindent\textbf{Set and Parameters}\vspace{0.3cm}\\
 \begin{tabular}{p{0.2cm} r p{15cm}}
 $\mathcal{J}^l$    & : & index set of the logarithm of the explanatory variables, $\mathcal{J}^l= \{m+1,...,2m\}$\\
 $\boldsymbol{A}^l$       & : & standardized data matrix formed of the logarithm of the explanatory variables, $\boldsymbol{A}^l= \lbrack a^l_{ij} \rbrack  \in \mathbb{R}^{n\times m}$ \\
 $\widetilde{\boldsymbol{A}}$       & : &  data matrix concatenating $\boldsymbol{A}$ and $\boldsymbol{A}^l$, i.e., $\widetilde{\boldsymbol{A}}=\lbrack \boldsymbol{A} \ \boldsymbol{A}^l \rbrack = \lbrack \tilde{a}_{ij} \rbrack \in\mathbb{R}^{n \times 2m}$
 \end{tabular}
 \vspace{0.2cm}

  If a vector corresponding to the $j$-th column of $\boldsymbol{A}$ possesses non-positive elements, we compute the logarithm of the column using a conventional method to deal with the non-positives: $a^l_{ij} = \log(a_{ij}+|\min(\lbrace a_{ij} | i\in \mathcal{I} \rbrace)|+1)$. Because all original explanatory variables have transformed variables, we can define $j^l = j + m$ for each $j^l \in \mathcal{J}^l$. The model incorporating the log-transformed explanatory variables, $\mathcal{MP}_\text{base}(k)$, can be formulated as follows:
 \begin{subequations}
 \begin{align}
 \mathcal{MP}_\text{base}(k):
 \quad \text{minimize }		& \quad \sum_{i\in \mathcal{I}}{ \left( {e_{i}^{+}}^2 + {e_{i}^{-}}^2 \right)} \label{f:3_0} \\
                               \text{subject to }	& \quad e_i^+ - e_i^- = \sum_{j\in \mathcal{J}\cup \mathcal{J}^l} \tilde{a}_{ij}\left( x_{j}^+ - x_{j}^- \right) - b_{i}
                                                                                  \> && \forall i \in \mathcal{I}, \label{f:3_1} \\
       						& \quad x_{j}^+ + x_{j}^- \leq M z_{j}, 			 \> && \forall j \in \mathcal{J} \cup \mathcal{J}^l, \label{f:3_2} \\
                             & \quad \sum_{j\in \mathcal{J} \cup \mathcal{J}^l} z_{j} = k,			 \> &&  \label{f:3_3} \\
                             & \quad z_{j} + z_{j^l} \leq 1, 						 \> && {\begin{aligned}
                                                                                         \forall & j \in \mathcal{J}, \> j^l=j+m,
                                                                                         \end{aligned}} \label{f:3_4} \\
                             & \quad x_{j}^+, x_{j}^-  \geq 0, 	 \> && \forall j \in \mathcal{J} \cup \mathcal{J}^l, \label{f:3_5}\\
                             & \quad e_{i}^+, e_{i}^-  \geq 0,	\> &&	\forall i \in \mathcal{I}, \label{f:3_7}\\
                             & \quad z_{j} \in \{ 0, 1 \},	\> &&	\forall j \in \mathcal{J} \cup \mathcal{J}^l. \label{f:3_8}
 \end{align}\label{f:3}
 \end{subequations}
 The extended model $\mathcal{MP}_\text{base}(k)$ is similar to formulation \eqref{f:2}. The difference is that \eqref{f:3} has additional derived explanatory variables in $\mathcal{J}^l$ and \eqref{f:3_4} ensures that both the original and log transformed explanatory variables are not selected at the same time. We note that, in addition to the log-transformation, other types of transformation of the explanatory variables can be considered in a similar manner. Finally, we remark that an appropriate value of $M$ in \eqref{f:3_2} is required to solve the problem. When $M$ is too small, we cannot guarantee optimality. When $M$ is too large, the optimization solver can struggle due to numerical issues. \citet{park2013subset} proposed a sampling-based approach, where $M$ is estimated by iteratively sampling a subset of explanatory variables. We use this method with a slight modification to ensure that the sampled explanatory variables do not simultaneously include an original explanatory variable and its transformation.

 \subsection{Inclusion of constraints corresponding to $t$-tests for the significance of the regression coefficients}\label{S:3_3}
 In this section, we present constraints to check the statistical significance of the regression coefficients. Formulating these constraints is not a trivial task because the statistical significance of an estimated coefficient depends on the selected subset. In detail, $s(A)_j$ in \eqref{eq:ttest_1} can only be calculated given a subset so that an inequality including $s(A)_j$ cannot be trivially formulated as a convex constraint. In order to address this issue, we convert the inequality \eqref{eq:ttest_2} into a constraint that checks the statistical significance of the $j$-th variable if it is selected. We derive
 \begin{subequations}\label{ineq:6}
 \begin{align}
 \> &\left| \frac{x_j}{s(\boldsymbol{A}_{s(k)})_j} \right| \geq t_{1-\frac{\alpha}{2} ,n-k-1} \label{ineq:6a}\\
 & \Longrightarrow \> \quad \left| x_j \right| 	 \geq t_{1-\frac{\alpha}{2} ,n-k-1} \left| s(\boldsymbol{A}_{s(k)})_j \right| - M(1-z_j)\label{ineq:6b}\\
 & \Longrightarrow \> \quad x^+_j + x^-_j 	\geq t_{1-\frac{\alpha}{2} ,n-k-1} s(\boldsymbol{A}_{s(k)})_j - M(1-z_j),\label{ineq:6c}
 \end{align}
 \end{subequations}
\noindent where $s(k)$ is a subset with $k$ explanatory variables, $\boldsymbol{A}_{s(k)}$ is a submatrix of $\boldsymbol{A}$ derived from $s(k)$, and $s(\boldsymbol{A}_{s(k)})_j$ is the standard deviation of the estimated coefficient for explanatory variable $j$ which is equivalent to \mbox{ $\sqrt{ \text{MSE}_{\boldsymbol{A}_{s(k)}}(\boldsymbol{A}_{s(k)}^\prime\boldsymbol{A}_{s(k)})^{-1}_j}$} and only defined for explanatory variables in $s(k)$ (i.e., $j \in s(k)$, but we omit the notation $j$ from $s(k)$ for notational simplicity). The resulting constraint above is only activated when the $j$-th variable is selected (i.e., $z_j = 1$). Note that $\boldsymbol{A}_{s(k)}$ varies according to the selected variables. It implies that the values $s(\boldsymbol{A}_{s(k)})_j$ in the constraint above vary accordingly and we cannot calculate the $s(\boldsymbol{A}_{s(k)})_j$ value if subset $s(k)$ is not fixed. Hence, it is still difficult to add constraint \eqref{ineq:6c} in its current form to the MIQP model.

 To handle changing values of $s(\boldsymbol{A}_{s(k)})_j$ depending on the selected subset, we use the lower bound of $s(\boldsymbol{A}_{s(k)})_j$, which gives a relaxed version of the exact constraint. A tight lower bound for $s(\boldsymbol{A}_{s(k)})_j$ can cut some of the subsets with insignificant regression coefficients. Also, the lower bound should be calculated very efficiently to avoid additional computational cost. 
 
 Let $R^2(\boldsymbol{A}, \boldsymbol{b})$ denote the coefficient of determination for a multiple linear regression model fitted to data matrices $\boldsymbol{A}$ and $\boldsymbol{b}$. To obtain the lower bound for $s(\boldsymbol{A}_{s(k)})_j$, we start with the following definition.
 \begin{definition}\label{def:1}
 In a multiple linear regression with design matrix $\boldsymbol{A}$, the variance inflation factor (VIF) of the $j$-th explanatory variable, denoted by $VIF_j$, is given as $VIF_j = \frac{1}{1-R^2(\boldsymbol{A}_{-j}, \boldsymbol{A}_{j})}$ where $\boldsymbol{A}_{-j}$ represents a submatrix of $\boldsymbol{A}$ formed by excluding the $j$-th column.
 \end{definition}
 Note that $R^2(\boldsymbol{A}_{-j}, \boldsymbol{A}_{j})$ is the coefficient of determination for a linear regression model, where variable $j$ is the response variable and all variables except for $j$ are explanatory variables. The quantity $VIF_j$ is indeed given by the $j$-th diagonal elements of $(\boldsymbol{A}^\prime\boldsymbol{A})^{-1}$. Now we recall the lemma from \citet{rencher2008linear}.
 \begin{lemma} \label{lemma:1}
    In a multiple linear regression model, if $s \subset s'$, then $R^2(\boldsymbol{A}_{s}, \boldsymbol{b}) \leq R^2(\boldsymbol{A}_{s'}, \boldsymbol{b})$. 
 \end{lemma}
 
  Lemma \ref{lemma:1} states that the coefficient of determination does not decrease by adding a new explanatory variable.   Based on the lemma, we derive the lower bound for $s(\boldsymbol{A}_{s(k)})_j$ to obtain the relaxed constraints.
 \begin{theorem}
 Let $R^2_{-j} =\underset{j'\in (\mathcal{J}\cup \mathcal{J}^l) \textrm{\textbackslash} \{j\}}{\min} R^2(\boldsymbol{A}_{j'}, \boldsymbol{A}_{j})$ for $j \in \mathcal{J}\cup \mathcal{J}^l$ and $k \geq 2$. Then for each $j$, $s(\boldsymbol{A}_{s(k)})_j \geq \sqrt{ \frac{\emph{MSE}_\emph{LB}(k)}{1-R^2_{-j}} },$
 where $\emph{MSE}_\emph{LB}(k)$ is the optimal solution of the LP-relaxation of  $\mathcal{MP}_\text{base}(k)$. 
 \end{theorem}
 
 \begin{proof}
    Because $s(\boldsymbol{A}_{s(k)})_j = \sqrt{\text{MSE}_{\boldsymbol{A}_{s(k)}}(\boldsymbol{A}_{s(k)}^\prime\boldsymbol{A}_{s(k)})^{-1}_j}$, we can calculate the lower bound of $s(\boldsymbol{A}_{s(k)})_j$ by obtaining lower bounds of $\text{MSE}_{\boldsymbol{A}_{s(k)}}$ and  $(\boldsymbol{A}_{s(k)}^\prime\boldsymbol{A}_{s(k)})^{-1}_j$.
    
    We first derive a lower bound of  $\text{MSE}_{\boldsymbol{A}_{s(k)}}$. Observe that $\text{MSE}_{\boldsymbol{A}_{s(k)}}$ corresponds to the optimal solution of $\mathcal{MP}_\text{base}(k)$.  Because the optimal value of a mixed integer programming model minimizing the objective function is lower bounded by that of its LP-relaxation, we have $\text{MSE}_{\boldsymbol{A}_{s(k)}} \geq \text{MSE}_{\text{LB}}$. 
    
    Now let us derive a lower bound of  $(\boldsymbol{A}_{s(k)}^\prime\boldsymbol{A}_{s(k)})^{-1}_j$. We start with noting that for an arbitrary $j$ and a subset $s(k)$ that includes $j$, there exists $j_0\in (\mathcal{J}\cup \mathcal{J}^l) \text{\textbackslash} \{j\}$ where $\{j_0\} \subset s(k)\text{\textbackslash}\{j\}$. Thus, by Lemma \ref{lemma:1} we can realize the inequalities  $R^2\left(\boldsymbol{A}_{s(k)\text{\textbackslash}\{j\}}, \boldsymbol{A}_{j}\right) \geq R^2(\boldsymbol{A}_{j_0}, \boldsymbol{A}_{j})\geq \underset{j'\in (\mathcal{J}\cup \mathcal{J}^l) \textrm{\textbackslash} \{j\}}{\min} R^2(\boldsymbol{A}_{j'}, \boldsymbol{A}_{j})=R^2_{-j}$. 
    Then, by Definition \ref{def:1} and the fact that $ 0 \leq R^2(\boldsymbol{A}, \boldsymbol{b}) \leq 1$ for any $\boldsymbol{A}$ and $\boldsymbol{b}$, we have
    \begin{equation*}
        (\boldsymbol{A}_{s(k)}^\prime\boldsymbol{A}_{s(k)})^{-1}_j = \frac{1}{1-R^2\left(\boldsymbol{A}_{s(k)\text{\textbackslash}\{j\}}, \boldsymbol{A}_{j}\right)} \geq \frac{1}{1-R^2_{-j}}.
    \end{equation*}
    
    Finally, combining the lower bounds of $\text{MSE}_{\boldsymbol{A}_{s(k)}}$ and $(\boldsymbol{A}_{s(k)}^\prime\boldsymbol{A}_{s(k)})^{-1}_j$, the lowerbound of $s(\boldsymbol{A}_{s(k)})_j$ is given by
    \begin{equation*}\pushQED{\qed} 
     s(\boldsymbol{A}_{s(k)})_j = \sqrt{\text{MSE}_{\boldsymbol{A}_{s(k)}}(\boldsymbol{A}_{s(k)}^\prime\boldsymbol{A}_{s(k)})^{-1}_j} \geq \sqrt{ \text{MSE}_\text{LB}(k)} \sqrt{\frac{1}{1-R^2_{-j}} }.\qedhere
    \end{equation*}
\end{proof}
 
 Setting $s_j(k)_{\text{LB}} = \sqrt{ \frac{\text{MSE}_\text{LB}(k)}{1-R^2_{-j}} }$, we can formulate the relaxed constraint of \eqref{ineq:6} as  
 \begin{equation}\label{eq:2}
 x^+_j + x^-_j 	\geq t_{1-\frac{\alpha}{2} ,n-k-1}{s_j(k)_{\text{LB}}} z_j, \qquad \forall j\in \mathcal{J}\cup \mathcal{J}^l .
 \end{equation}
 
It is worthy to note that calculating $s_j(k)_{\text{LB}}$ is computationally efficient. The quantity $\text{MSE}_\text{LB}(k)$ is obtained by solving an LP and the coefficient of determination $R^2_{-j}$ can be quickly obtained by the closed-form formula for $(m-1)$ simple linear regression models.
 
  Because $s_j(k)_{\text{LB}}$ is a calculated constant and $t_{1-\frac{\alpha}{2} ,n-k-1}$ is determined by the user, constraint \eqref{eq:2} is a linear constraint. However, constraint \eqref{eq:2} does not work properly with \eqref{f:3}. This is because, given a solution to \eqref{f:3}, both $x_j^+$ and $x_j^-$ can be increased to satisfy \eqref{eq:2} without violating the other constraints in \eqref{f:3} or changing the objective function value of the solution. To handle this issue, we introduce new variables $z_j^+$ and $z_j^-$, where $z_j^+ = 1$ if $x_j^+ > 0$, $z_j^+=0$ if $x_j^+ = 0$ and $z_j^- = 1$ if $x_j^->0$, $z_j^- = 0$ if $x_j^- = 0$. By replacing $z_j$ with $z_j^+$ and $z_j^-$, \eqref{f:3} can be modified as follows:
  \begin{subequations}
 	\begin{align}
 	  \text{minimize }		& \quad \sum_{i\in \mathcal{I}}{ \left( {e_{i}^{+}}^2 + {e_{i}^{-}}^2 \right)} \label{f:4_0} \\
 	\text{subject to }	& \quad e_i^+ - e_i^- = \sum_{j\in \mathcal{J}\cup \mathcal{J}^l} \tilde{a}_{ij}\left( x_{j}^+ - x_{j}^- \right) - b_{i}
 	\> && \forall i \in \mathcal{I}, \label{f:4_1} \\
 	& \quad x_{j}^+ \leq M z^+_{j}, 			 \>\> && \forall j \in \mathcal{J} \cup \mathcal{J}^l, \label{f:4_2} \\
 	& \quad x_{j}^- \leq M z^-_{j},				\>\> && \forall j \in \mathcal{J} \cup \mathcal{J}^l, \label{f:4_3} \\
 	& \quad \sum_{j\in \mathcal{J} \cup \mathcal{J}^l} \left( z^+_{j} + z^-_{j} \right) = k,			 \>\> &&  \label{f:4_4} \\
 	& \quad z^+_{j} + z^-_{j} + z^+_{j^l} + z^-_{j^l} \leq 1, 						 \>\> && {\begin{aligned}
 		\forall & j \in \mathcal{J}, \>\> j^l=j+m,
 		\end{aligned}} \label{f:4_5} \\
 	& \quad x^+_j + x^-_j 	\geq t_{1-\frac{\alpha}{2} ,n-k-1}{s_j(k)_{\text{LB}}} \left(z^+_j+z^-_j\right), \>\> && \forall j\in \mathcal{J}\cup \mathcal{J}^l, \label{f:4_6}\\
 	& \quad x_{j}^+, x_{j}^-  \geq 0, \quad z^+_{j}, z^-_{j} \in \{ 0, 1 \},	 \>\> && \forall j \in \mathcal{J} \cup \mathcal{J}^l, \label{f:4_7}\\
 	& \quad e_{i}^+, e_{i}^-  \geq 0,	\>\> &&	\forall i \in \mathcal{I}. \label{f:4_8}
 	\end{align}\label{f:4}
 \end{subequations}
 Note that $z_j$ is replaced by $z_j^+$ or $z_j^-$ in \eqref{f:4_2} and \eqref{f:4_3} and $z_j^+ + z_j^-$ is used instead of $z_j$ in \eqref{f:4_4} and \eqref{f:4_5}. Note that \eqref{f:4_6} is added for the $t$-test, but it is a relaxed version of the exact $t$-test constraint \eqref{ineq:6}. This relaxed constraint is useful in reducing the search space of the branch-and-bound algorithm of the solver, while the exact constraint is enforced by the technique presented in the subsequent section.

 \subsection{Final algorithm}\label{S:3_4}
  Recall that constraint \eqref{f:4_6} is a relaxed constraint derived from the lower bound of $s(A_{s(k)})$ and that a feasible solution for \eqref{f:4} can have statistically insignificant coefficients, while we want all coefficients to be statistically significant. Furthermore, it is challenging to formulate tests to check the residual assumptions in Section \ref{S:2_3} as linear or convex constraints. To overcome these limitations, we propose a lazy constraint-based approach to solve the following problem.
 
 \begin{equation}\label{f:5}
 	\begin{aligned} 
 	\quad \text{minimize }		& \quad \eqref{f:4_0} \\
 	\text{subject to }			& \quad \eqref{f:4_1} \text{-} \eqref{f:4_8}, \\ 
 								& \quad \text{coefficients $t$-tests constraints \eqref{eq:ttest_2} from Section \ref{S:2_2},} \\
 								& \quad \text{residual diagnostics constraints from Section \ref{S:2_3}.}
 	\end{aligned}
 \end{equation}
 
 Lazy callback is mainly employed in practical implementations using an optimization solver when the number of constraints in an MIP model is extremely large \citep{gurobi}. For example, we may use lazy callback when solving the traveling salesman problem (TSP) because it has a large number of subtour elimination constraints. Instead of adding all subtour elimination constraints at the root node of the branch-and-bound algorithm, we can iteratively and selectively add some of these constraints. In detail, when the solver arrives at a branch-and-bound node and the solution contains a subtour, lazy callback allows us to add the corresponding subtour elimination constraint at the current node. Although our MIQP model does not have an extremely large number of constraints, we can use lazy callback to stop at a branch-and-bound node and check if the solution completely satisfies all of the tests and diagnostics. Recently, \citet{Bertsimas2019Accounting} also use the lazy constraint to overcome difficulties related to the coefficient $t$-test. Our lazy constraint-based procedure, conducted independently of
 \citet{Bertsimas2019Accounting}, is provided in Algorithm \ref{alg:lazy}.

 \begin{algorithm}[h!]
     \caption{Lazy constraint procedure (at each node ${J}_{\text{node}}$)}\label{alg:lazy}
     \begin{algorithmic}[1]
     \INPUT a set of $\alpha$ (significance levels for each statistical test)
     \FOR {\text{each} $j' \in {J}_{\text{node}}$}
         \IF {the coefficient corresponding to variable $j'$ is not statistically significant}
       	     \STATE add a lazy constraint $\sum_{j\in{{J}_{node}}}\left(z^+_{j} + z^-_{j}\right) \leq k-1$ to the model
 		     \STATE \textbf{return}
         \ENDIF
     \ENDFOR
     \IF {the model with variables in ${J}_{\text{node}}$ fails to pass the residual diagnostics test}
     \STATE add a lazy constraint $\sum_{j\in{J_{node}}}\left(z^+_{j} + z^-_{j}\right) \leq k-1$ to the model
     \ENDIF
     \STATE \textbf{return}
   \end{algorithmic}
 \end{algorithm}

 In the algorithm, $J_{node}$ is the index set of selected variables at the current node in the brand-and-bound tree. The lazy constraint $\sum_{j\in{J_{node}}}\left(z^+_{j}+z^-_{j}\right) \leq k-1$ is added to the model when (i) a linear model associated with $j\in J_{node}$ fails to pass at least one of the statistical tests for the coefficients or (ii) the model fails to pass the residual diagnostics test. The lazy constraint eliminates the solution of selecting all explanatory variables in $J_{node}$ from the feasible region. For example, let us consider a problem with 10 explanatory variables with the associated binary variables set $\lbrace{z_1,...,z_{10}}\rbrace$. We assume that we set $k=4$ and the solver is currently at the node with a solution having $z^+_1= z^+_4 = z^-_9 = z^+_{10} = 1$ and $z^+_j, z^-_j = 0$ for all others. The procedure generates a linear model with the selected explanatory variable set $J_{node}= \lbrace 1,4,9,10 \rbrace$ and conducts statistical tests and diagnostics for the model. When the regression model fails to pass at least one of the tests, lazy constraint $\sum_{j\in{J_{node}}}\left(z^+_{j}+z^-_{j}\right) = z^+_1 + z^+_4+ z^-_9 + z^+_{10} \leq 3$ is added to the MIQP model. As a result, the solution $z^+_1= z^+_4 = z^-_9 = z^+_{10} = 1$ (the others are zero due to constraint \eqref{f:4_4}) becomes infeasible in all of the subsequent branches. 
 
 When there is no feasible solution with a fixed $k$ value, the user can relax the cardinality constraint and consider different $k$ values. This search can easily be performed by re-running the algorithm with different $k$ values; we do not consider multiple $k$ values in the proposed alternative solution procedure. 
 
 Table \ref{tab:modelsummary} summarizes the model building and validation procedures presented in this section, along with the associated constraints and sections.  Finally, we remark that the proposed lazy constraint framework can easily be extended and incorporate other tests such as \textit{F}-tests or pairwise multicollinearity tests based on user needs.

 \begin{table}[htbp]
  \centering
  \caption{Constraints for model building and validation procedures}
    \adjustbox{max width=0.8\textwidth}{ \begin{tabular}{|c|c|c|}
    \hline
    Model building and validation procedure & Associated constraints & Section reference \\
    \hline
    Nonlinear variable transformations & \eqref{f:4_5}      & Section \ref{S:2_1} \\
    \hline
    Significance of estimated coefficients & \eqref{f:4_6}, lazy constraint      & Section \ref{S:2_2}\\
    \hline
    Residual tests &  lazy constraint     & Section \ref{S:2_3}\\
    \hline
    \end{tabular}}%
  \label{tab:modelsummary}%
\end{table}%
 
 \section{Alternative solution procedure for infeasible problems}\label{S:4}
 
  Based on the algorithm in Section \ref{S:3_4}, we can obtain the best subset satisfying all tests for model validation discussed in Section \ref{S:2}. However, when all possible subsets violate at least one of the tests and assumptions, the model \eqref{f:4} becomes infeasible and the algorithm will not return a solution. Even in this case, a near-feasible solution is desired (i.e., satisfying most of the tests and assumptions while mildly violating a few of the tests). To find such solutions, we propose an algorithm, referred to as \textit{alternative solution procedure}. The algorithm searches for a near-feasible solution based on the model development strategy in Figure \ref{fig:0} and penalties for violation of regression tests and assumptions. 
  
  The alternative solution procedure is invoked at each branch-and-bound node while no feasible solution satisfying all tests and diagnostics has been found by the algorithm. Once a feasible solution is found, the alternative solution procedure will not be invoked because at least one feasible solution to the problem satisfying all of the tests and diagnostics exists. When the alternative solution procedure is invoked, it compares a \emph{new subset} with the incumbent \emph{best subset} (not feasible, but the best alternative near-feasible subset) and decides which subset is better. Our model updates and keeps only one alternative solution until it finds a feasible solution. The new subset, denoted as $S_\text{new}$, is the subset at the current branch-and-bound node and the incumbent best subset, denoted as $S_\text{best}$, is the best alternative subset found so far. Remark that $S_{\text{new}}$ and $S_{\text{best}}$ must be infeasible solutions with some insignificant coefficients or residual assumption violations because the alternative solution procedure keeps $S_\text{best}$ until a feasible solution is found. To simplify the discussion, we will use $S_\text{best}$ and $S_\text{new}$ to refer to both selected subsets and linear regression models.
 
  Note that one of the simplest approaches to comparing $S_\text{new}$, $S_\text{best}$, and all other subsets is to add penalty terms to the objective function based on diagnostic violations such as the average $p$-value or the number of insignificant variables. Before presenting the details of the alternative solution procedure, we first explain why the simple penalty approach may fail and why the proposed procedure is needed. First, the penalty approach cannot effectively reflect the overall development procedure of a linear regression model. In typical model development, explained in Section \ref{S:2}, we first establish a linear regression model with statistically significant coefficients. We then check the residual assumptions via diagnostics. That is, we should consider the residual assumptions after establishing a statistically significant model. This cannot be achieved with the simple penalty approach. Next, there exist non-trivial cases where the penalty approach may fail to select the better subset. This problem is discussed with the illustrative examples presented in Table \ref{table:Alt_example}.
 
 \begin{table}[h!]  	  \caption{Illustrative examples for the alternative solution procedure}
 	\begin{center}
 		\noindent\adjustbox{max width=0.7\textwidth}{
 		\begin{tabular}{|c| c l |c|}
 			\hline
 			\multicolumn{1}{|c|}{Cases} & \multicolumn{2}{c|}{Set of $p$-values} &  \multicolumn{1}{c|}{Better set} \\
 			\hline\rule{0pt}{2.5ex}
 			\multirow{2}{*}{Case 1} & \hspace{0.3cm} $S_{\text{best}}$ & $\lbrace \mathbf{0.06}, \mathbf{0.06}, \mathbf{0.06}, 0.025, 0.02 \rbrace$ & \multirow{2}{*}{$S_{\text{new}}$} \\
 			& \hspace{0.3cm} $S_{\text{new}}$ & $\lbrace \mathbf{0.055}, \mathbf{0.055}, 0.04, 0.025, 0.02 \rbrace$ & \\
 			
 			\hline\rule{0pt}{2.5ex}
 			\multirow{2}{*}{Case 2} & \hspace{0.3cm} $S_{\text{best}}$ & $\lbrace \mathbf{0.06}, \mathbf{0.06}, \mathbf{0.06}, 0.025, 0.02 \rbrace$ & \multirow{2}{*}{$S_{\text{best}}$}\\
 			& \hspace{0.3cm} $S_{\text{new}}$ & $\lbrace \mathbf{0.95}, 0.04, 0.04, 0.025, 0.02 \rbrace$ & \\
 			\hline\rule{0pt}{2.5ex}
 			\multirow{2}{*}{Case 3} & \hspace{0.3cm} $S_{\text{best}}$ & $\lbrace \mathbf{0.06}, \mathbf{0.06}, \mathbf{0.06}, 0.025, 0.02 \rbrace$ & \multirow{2}{*}{$S_{\text{new}}$}\\
 			& \hspace{0.3cm} $S_{\text{new}}$ & $\lbrace \mathbf{0.065}, 0.04, 0.04, 0.025, 0.02 \rbrace$ & \\
 			\hline
 		\end{tabular}}
 	\label{table:Alt_example}
 	\end{center}
 \end{table}
 
 Table \ref{table:Alt_example} presents three cases of determining which solution is better between $S_{\text{best}}$ and $S_{\text{new}}$, where $k$ and $\alpha$ are set to 5 and 0.05, respectively. The set of $p$-values from the statistical test for the coefficients is shown for each case. In Case $1$, it is not surprising that we choose $S_{\text{new}}$ because the number of insignificant coefficients in $S_{\text{new}}$ is less than that in $S_{\text{best}}$ (i.e., $3>2$), and the average of the violating $p$-values for $S_{\text{new}}$ is also less than that for $S_{\text{best}}$ (i.e., $0.060>0.055$). On the other hand, $S_{\text{best}}$ must be better in Case 2 because the average of the violating $p$-values for $S_{\text{new}}$ is much larger than that for $S_{\text{best}}$ ($0.06<0.95$), even though the number of insignificant coefficients in $S_{\text{new}}$ is greater than that in $S_{\text{best}}$ ($3>1$). In Case 3, although the average of the violating $p$-values for $S_{\text{best}}$ is less than that for $S_{\text{new}}$ ($0.60<0.65$), selecting $S_{\text{new}}$ as the better solution seems reasonable because the difference is negligible while the number of insignificant coefficients in $S_{\text{best}}$ is significantly greater than in $S_{\text{new}}$ (i.e., $3>1$). These illustrative examples indicate that we should consider both the average of the violating $p$-values and the number of insignificant coefficients.  However, due to the different magnitudes of the measures, it is not trivial to set appropriate weights. Consequently, it is necessary to develop an intuitive procedure that reflects the framework of linear regression model construction and further alleviates the scale difference for measures of diagnostic violation. The experiment in Section \ref{S:5} shows that the alternative solution procedure outperforms a simple penalty approach.
 
  Now, we discuss the alternative solution procedure. To explain this more efficiently, we first define the functions related to the significance test for the coefficients.
 
 \noindent\begin{longtable}{l r p{13.2cm}}
 	$\pi(S)$       				& : & number of insignificant coefficients in the model with subset $S$ \\
 	$E(S)$       				& : & average $p$-values of the insignificant coefficients in the model with subset $S$; $0$ if all coefficients in subset $S$ are statistically significant\\
 	$r_l(S)$       				& : & $p$-value from the residual linearity test for subset $S$ \\
 	$r_h(S)$       				& : & $p$-value for the residual heteroscedasticity tests, which is the maximum $p$-value between the $p$-values of the absolute residual test and the Breusch-Pagan test\\
 	 	$f_q(S_1, S_2)$       		& : & \textit{decision by quality}; $S_1$ and $S_2$ are subsets  \\
 	$f_t(S_1, S_2; \tau)$		& : & \textit{decision with tolerance}; $S_1$ and $S_2$ are subsets \\
 \end{longtable}

  The last two functions, $f_q(S_1, S_2)$ and $f_t(S_1, S_2; \tau)$, decide whether $S_1$ or $S_2$ is better based on different principles, which will be discussed in detail later in this section. 
 
 The overall alternative solution procedure is presented in Algorithm \ref{alg:asp}. Step 1 is for when $S_{\text{new}}$ is the statistically significant model while $S_{\text{best}}$ is not. Hence, it is natural to return $S_{\text{new}}$. Step 2 considers the opposite case to that of Step 1. In Steps 3-4, non-trivial cases are considered: both models are statistically significant in Step 3 and both models are statistically insignificant in Step 4. For these two cases, $f_q$, referred to as decision by quality, and $f_t$, referred to as decision by tolerance, are employed to make a decision.
 
 \begin{algorithm}[h!]
 	\caption{Alternative solution procedure}\label{alg:asp}
 	\begin{algorithmic}[1]
 		\INPUT $S_{\text{best}}$, $S_{\text{new}}$
 		\OUTPUT the better subset between $S_{\text{best}}$ and $S_{\text{new}}$
 		\IF {$ \pi(S_{\text{best}}) > 0 $ \textbf{and} $ \pi(S_{\text{new}}) = 0 $} \textbf{return} $S_{\text{new}}$
 		\ELSIF {$ \pi(S_{\text{best}}) = 0 $ \textbf{and} $ \pi(S_{\text{new}}) > 0 $} \textbf{return} $S_{\text{best}}$
 		\ELSIF {$ \pi(S_{\text{best}}) = 0 $ \textbf{and} $ \pi(S_{\text{new}}) = 0 $} \textbf{return} $f_q(S_{\text{best}}, S_{\text{new}})$
 		\ELSIF {$ \pi(S_{\text{best}}) > 0 $ \textbf{and} $ \pi(S_{\text{new}}) > 0 $} \textbf{return} $f_t(S_{\text{best}}, S_{\text{new}}; \tau)$
 		\ENDIF
 	\end{algorithmic}
 \end{algorithm}
 
 Function $f_q(S_{\text{best}}, S_{\text{new}})$ returns the better solution between $S_{\text{best}}$ and $S_{\text{new}}$ by quantifying the quality of the solutions using the penalty function $q$ for solution $S$, introduced below.
 \vspace{0.5cm}
 
 \begin{tabular}{p{1.9cm} r p{13cm}}
 	$\text{MSE}(S)$       		& : & the MSE of solution $S$ \\
 	$\lambda_\pi, \lambda_E, \lambda_l, \lambda_h$ & : & penalty parameters for $\pi$, $E$, $r_l$, and $r_h$, respectively\\
 	$w_1(p, \alpha)$			& : & percentage transform function for $E(S)$; $\frac{\max{(p-(1-\alpha), 0)}}{\alpha}$ \\
 	$w_2(p, \alpha)$			& : & percentage transform function for $r_l(S)$ and $r_h(S)$; $\frac{\max((1-\alpha)-p, 0)}{1-\alpha}$
 \end{tabular}
 
  \begin{equation*}
 q(S; \lambda_\pi, \lambda_E, \lambda_{l}, \lambda_{h}) = \text{MSE}(S) + \lambda_\pi \pi(S) + \lambda_E w_1(E(S), \alpha_{E}) + \lambda_{l} w_2(r_l(S), \alpha_{l}) + \lambda_h w_2(r_{h}(S), \alpha_{h})
 \end{equation*} 
 Note that function $q$ includes percentage transform functions $w_1(p,\alpha)$ and $w_2(p,\alpha)$ to indicate the percentage gap between the significance levels. We introduce these functions because first we want to accurately measure the insignificance of the $p$-values when the significance levels are at different scales, and second, we do not want to apply penalties if the $p$-value falls within the significance level.
 
Two examples are provided for $w_1$ and $w_2$ to support our claims and demonstrate their necessity. The first example is of penalty nullification. Suppose $r_h(S_1)=0.2$ and $r_h(S_2)=0.04$, which implies the residuals of $S_1$ are consistent while those of $S_2$ have heteroscedasticity. Given $\alpha_h=0.9$, we can calculate $w_2(r_h(S_1), \alpha_{h}) = \frac{\max((1-0.9)-0.2, 0)}{0.1} = 0$, and $w_2(r_h(S_2), \alpha_h) = \frac{\max((1-0.9)-0.04, 0)}{0.1} = \frac{0.06}{0.1} = 0.6$. The penalties make sense because $S_1$ does not violate the test. Hence, the evaluation of $S_1$ is based only on the MSE and the significance of the coefficients. The second example demonstrates the role of scaling between $p$-values from different statistical tests. Suppose that we get $E(S_3) = 0.15$, $r_{h}(S_3) = 0.05$. Further, $\alpha_{E}$ and $\alpha_{h}$ are both set to $0.9$. Then, $S_3$ violates the $t$-tests of the coefficients by $0.15 - (1-0.9) = 0.05$, which is equal to the value of the heteroscedasticity test violation $(1-0.9) - 0.05 = 0.05$. However, the violation of $0.05$ is relatively small for the possible $t$-test violation range $[0,0.9]$ compared to the  possible residual test violation range $[0,0.1].$ Thus, despite the same violation size, the significance of the violation can differ depending on the tests and diagnostics. We can scale the magnitude of the violations through $w_1$ and $w_2$: $w_1(E(S_3), \alpha_{E}) = \frac{0.15-0.1}{0.9} \approx 0.056$, and  $w_2(r_{h}(S_3), \alpha_{h}) = \frac{0.1-0.05}{0.1} = 0.5$.

 We now discuss the \textit{decision with tolerance} presented in Algorithm \ref{alg:dt}. Step 1 is key to our procedure. If $E(S_{\text{new}})-E(S_{\text{best}})$ exceeds the tolerance $\tau$, a positive parameter predetermined by the user, then the alternative solution procedure concludes that $S_{\text{best}}$ is better. When the average of the violating $p$-values of $S_\text{new}$ is smaller or not significantly worse (smaller than $\tau$) than that of $S_\text{best}$, we conclude $S_\text{new}$ is competitive and investigate further in Steps 3-7 by comparing the two solutions based on two factors, $\pi$ and $E$. Step 3 concludes that, if $S_{\text{new}}$ is smaller than $S_{\text{best}}$ in terms of these two factors, $S_\text{new}$ is better. The opposite case is shown in Step 4. If there is a conflict between the two factors, $f_q$ settles the decision. Note that Algorithms \ref{alg:asp} and \ref{alg:dt} imitate the linear regression model building procedure described in Section 2 by using statistical significance tests first to make the decision, followed by other diagnostics. 
 
  \begin{algorithm}[h!]
 	\caption{\textit{Decision with tolerance, $f_t$}}\label{alg:dt}
 	\begin{algorithmic}[1]
 		\INPUT $S_{\text{best}}$, $S_{\text{new}}$, $\tau$
 		\OUTPUT the better subset between $S_{\text{best}}$ and $S_{\text{new}}$
 		\IF {$E(S_{\text{new}})-E(S_{\text{best}}) > \tau$} \textbf{return} $S_{\text{best}}$
 		\ELSE
 		\IF {$E(S_{\text{best}})>E(S_{\text{new}})$ \textbf{and} $\pi(S_{\text{best}})>\pi(S_{\text{new}})$} \textbf{return} $S_{\text{new}}$
 		\ELSIF {$E(S_{\text{best}}) < E(S_{\text{new}})$ \textbf{and} $\pi(S_{\text{best}}) < \pi(S_{\text{new}})$} \textbf{return} $S_{\text{best}}$
 		\ELSE \textbf{ return} $f_q(S_{\text{best}}, S_{\text{new}})$
 		\ENDIF
 		\ENDIF
 	\end{algorithmic}
 \end{algorithm}
 
  We illustrate the entire alternative solution procedure using the three toy examples presented in Table \ref{table:Alt_example}. Suppose that our model is searching a subset whose cardinality $k$ is set to 5 and that a feasible solution has not yet been found. A user sets the significance levels and $\tau$ to 95 percent and $0.1$, respectively.
 
 \begin{itemize}
 	\item \textbf{Case 1.}
 	Our model cannot determine which solution is better in the first step because $\pi(S_{\text{best}}) = 3 > 0$ and also $\pi(S_{\text{new}}) = 2 > 0$. Hence, decision with tolerance is invoked (Algorithm \ref{alg:dt}). In Step 1, the algorithm computes $E(S_{\text{best}}) = \frac{1}{3}(0.06+0.06+0.06) = 0.06$, $E(S_{\text{new}})= \frac{1}{2}(0.055+0.055) = 0.055$, and $E(S_{\text{new}}) - E(S_{\text{best}}) = -0.005 < \tau = 0.1$. Thus, the next steps compare $E(S_{\text{new}})$ with $E(S_{\text{best}})$ and $\pi(S_{\text{new}})$ with $\pi(S_{\text{best}})$. Since $E(S_{\text{new}}) < E(S_{\text{best}})$ and $\pi(S_{\text{new}}) < \pi(S_{\text{best}})$, $S_{\text{new}}$ is determined to be the better solution.
 	
 	\item \textbf{Case 2.}
 	Because $\pi(S_{\text{best}})$ and $\pi(S_{\text{new}})$ are both greater than $0$, our model cannot determine which solution is better in Step 4 of Algorithm \ref{alg:asp}, and $f_t$ is called. In Algorithm \ref{alg:dt}, since $E(S_{\text{new}}) - E(S_{\text{best}}) = 0.95 - 0.06 = 0.89 > \tau = 0.1$ in Step 1, $S_\text{best}$ is returned.
 	
 	\item \textbf{Case 3.}
 	Similar to the previous cases, $f_t$ is called in Step 4 of Algorithm \ref{alg:asp}. In Algorithm \ref{alg:dt}, although $E(S_{\text{new}}) - E(S_{\text{best}}) = 0.065 - 0.06 = 0.005$ and $E(S_{\text{new}})$ is greater than $E(S_{\text{best}})$, Steps 3-6 are considered because $E(S_{\text{new}}) - E(S_{\text{best}}) = 0.005 < 0.1 =\tau$. However, since $E(S_{\text{new}})$ is greater than $E(S_{\text{best}})$ $(0.065 > 0.06)$ and $\pi(S_{\text{new}}) $ is less than $\pi(S_{\text{best}}) $ $(1 < 3)$, the winning solution is determined using $f_q$.
 \end{itemize}

 Overall, we find that the alternative solution procedure selects the desired solution discussed in Table \ref{table:Alt_example}. During the alternative solution procedure, we first compare $S_\text{new}$ with $S_\text{best}$ using the significance of the estimated coefficients, in particular with $\pi(S)$ and $E(S)$. If this step cannot lead to a decision, residual diagnostics are considered. This multistage procedure reflects the linear regression model building framework discussed in Section \ref{S:2}. The procedure also deals with the possible issues illustrated in Case 2 and 3 in Table $\ref{table:Alt_example}$ by introducing $f_t$. A user can intuitively provide a value of $\tau$ based on the average difference in the $p$-value that they can permit. In Experiment 4 of Section \ref{S:5}, we provide experimental results that illustrate the benefit of using the proposed procedure rather than a simple penalty function.
 
 In this alternative solution framework, we empirically adopt $E(S)$ and $\pi(S)$ as measures evaluating how statistically significant a regression model is. The practical examples discussed above justify the measures. We finally note that these measures can be varied or alternatives can be employed to reflect their own perspectives on the statistical significance. In the next sections, $\mathcal{MP}_\text{lazy}(k)$ denotes our final algorithm: solving \eqref{f:4} with Algorithm \ref{alg:lazy} and the alternative solution procedure in Algorithms \ref{alg:asp} and \ref{alg:dt}. 

 Finally, we summarize the proposed approach as follows. Our ultimate goal is solving the MIP model \eqref{f:5}. Due to the nonlinearity of the exact $t$-test and diagnostic constraints, we instead start with solving the MIP model \eqref{f:4}. While solving the model, a lazy constraint for the $t$-test and diagnostic is added at each branch-and-bound node if needed. Simultaneously, we keep the best near-feasible solution found by the alternative solution procedure invoked at the node. Once a feasible solution is found during the search, the alternative solution procedure is not invoked. If we do not find a feasible solution until the solution search terminates, the model outputs the best near-feasible solution found by the alternative solution procedure.

\section{Computational Results} 
\label{S:5}
In this section, the results of numerical experiments using the proposed model and benchmarks are presented. The experiments are designed to demonstrate the performance of the proposed model and the necessity for the alternative solution procedure. 

\subsection{Benchmark dataset description and preprocessing}

Twelve publicly available datasets are used in the experiment. We collect five datasets for regression from the UCI machine learning data repository \citep{Lichman2013} and seven datasets from other sources. The features of the experimental datasets are summarized in Table \ref{Table:dataset}. 

\setcounter{table}{2}
\begin{table}[h!] 	\caption{Datasets used in the experimental demonstration}
		\begin{center}
		\noindent\adjustbox{max width=0.8\textwidth}{%
			\begin{tabular}{|c|c|c|c|c|}
				\hline
				\multirow{2}{*}{Dataset} &  
				\multirow{2}{*}{Observations ($n$)} &
				\multicolumn{2}{c|}{Explanatory variables} & \multirow{2}{*}{Source}
				\\
				\cline{3-4}
				& &Raw & Preprocessed ($m$)& \\
				\hline
				Housing 		& $506$ 	& 	$13$ 	&	$26$	&	UCI data repository\\
				Servo 			& $167$ 	& 	$4$ 	&	$38$	&	UCI data repository\\
				AutoMPG 		& $392$ 	& 	$8$ 	&	$50$	&	UCI data repository\\
				Automobiles 	& $159$ 	&	$26$ 	&	$124$	&	UCI data repository\\
				Winequality		& $1599$ 	& 	$11$ 	&	$22$	&	UCI data repository\\			
				Bodyfat 		& $252$ 	& 	$15$ 	&	$30$	&	\citet{johnson1996fitting}\\
				Barro 			& $161$ 	& 	$13$ 	&	$26$	&	\citet{koenker1999goodness}\\
				Carseats 		& $400$ 	& 	$10$ 	&	$28$	&	\citet{james2013introduction}\\
				Crime 			& $630$ 	& 	$22$ 	&	$50$	&	\citet{cornwell1994estimating}\\
				Framing 		& $265$ 	& 	$14$ 	&	$30$	&	\citet{brader2008triggers}\\
				Griliches 		& $758$ 	&	$19$ 	&	$50$	&	\citet{blackburn1991unobserved}\\
				Hprice3			& $321$		&	$19$	&	$22$	&	\citet{woodbridge2006introductory}\\
				\hline
		\end{tabular}}
	\end{center}
	\label{Table:dataset}
\end{table}

We preprocess the datasets as follows. First, we remove records which include at least one missing value. Second, dummy variables are introduced for the categorical variables. Third, nominal variables are changed into numerical variables. Next, log-transformed variables are created for the numerical variables in the original data. Finally, all of the datasets are standardized. 

\subsection{Experimental design}\label{S:5.2}

In this section, we present the design of the five experiments. To demonstrate the practical viability of our model, we set a time limit of 600 seconds for every experiment, except for Experiment 2 where the time limit is set to 3600 seconds to compare the performances when the algorithms have sufficient amount of time. The algorithm time includes the computation time for big-$M$ in \eqref{f:3_2} and $s(k)_\text{LB}$ in \eqref{eq:2}. We set several parameters for the experiments: $\alpha_E=0.95$ and $\alpha_{l}= \alpha_{h}=0.99$, where $\alpha_{E}$, $\alpha_{l}$, and $\alpha_{h}$ are the significance levels for the $t$-tests for the coefficients, diagnostics for residual linearity, and residual heteroscedasticity, respectively. Parameters for the alternative solution procedure is set to as follows: $\tau = 0.1; \lambda_E = 1.5; \lambda_\pi = \lambda_h = \lambda_l = 0.125$.  The parameters are tuned to balance the fitting errors and validation measures based on the datasets used in our experiment. In our pilot computational experiments, we found that the suggested parameters return good results for various datasets. Therefore, for a new dataset, we recommend the user to start with the suggested parameters and iteratively adjust the parameters to improve the specific validation measures $\pi, E, r_l,$ and $r_h$ as needed. Alternatively, one can choose the parameters by cross-validation. In general, increasing the penalty parameter improves the corresponding validation measure. In Experiment 5, we show how the validation measures change as we change the penalty parameters and we hope this will help the user to successfully search for good parameters and determine their own best parameters.

\noindent\textit{\textbf{Experiment 1: Proposed model vs. simple benchmark models}}
We compare $\mathcal{MP}_\text{lazy}(k)$ with three benchmark models: $\mathcal{MP}_\text{base}(k)$, $\mathcal{FS}(k)$ (a forward selection algorithm), and LASSO. With this experiment, the performance of the proposed model in terms of the aforementioned statistical significance and regression assumptions can be verified. 

The forward selection algorithm iteratively adds an explanatory variable until $k$ variables are selected while the original and log-transformed variables cannot be in the subset simultaneously. The algorithm is constructed by modifying a standard forward selection algorithm to select from among variables and their log-transforms. Specifically, once a variable is selected, both the original and transformed variables are excluded from the candidate set of variables. If $j^* \in J$, then the log-transformed variable is excluded as well. If $j^* \in J^l$, then the original variable is excluded as well. Thus, $\mathcal{FS}(k)$ always provides a solution, with at most one variable from an original and transformed variable pair.

LASSO is a very popular regularized regression method that penalizes the $L_1$ norm of coefficients. The key advantage of LASSO is that it automatically performs variable selection while the coefficients shrink through the regularization.  LASSO and the proposed model are different in that our model has an $L_0$ norm-based constraint (penalty) and diagnostic constraints. Intrinsically, it is not straightforward to include the additional constraints in LASSO.

For each dataset, all models and algorithms are tested over $k=3, 4,...,10$. Because LASSO does not select $k$ explicitly, the proper penalty parameters of LASSO that give $k = 3,4,...,10$ are found by grid search.  For each case, we check the goodness-of-fit with an adjusted $R^2$, denoted as $R^2_{adj}$. Although the objective of all models and algorithms is to maximize the SSE, this is equivalent to maximizing $R^2_{adj}$ because $k$ is fixed. Also, because $R^2_{adj}$ ranges from $0$ to $1$, we can easily compare the goodness-of-fit over different $k$ values and datasets.

\noindent\textit{\textbf{Experiment 2: Proposed model vs. iterative model}}
We compare our model with the iterative algorithm used in \citet{bertsimas2016OR} for cases where the solution for $\mathcal{MP}_\text{base}(k)$ violates some of the tests and diagnostics. The algorithm iteratively adds constraints to avoid subsets with insignificant coefficients. We will refer to the iterative model as $\mathcal{MP}_\text{iter}(k)$ in this paper. Recall that a key feature of our algorithm is the ability to avoid iterative calls of the MIQP solver, where iterative approaches solve multiple MIQPs by adding constraints. In this comparison, we demonstrate the effectiveness of our lazy constraint-based algorithm. As the two models have different constraints and parameters, we compare the two models using common constraints, the number of variables selected and $t$-tests. That is, we do not consider the residual tests in this experiment.

\noindent\textit{\textbf{Experiment 3: Proposed model vs. model without the relaxed $t$-test constraints}} We show the effectiveness of the relaxed $t$-test constraints \eqref{f:4_6} derived in Section \ref{S:3_3} by comparing our model with and without the relaxed constraint \eqref{f:4_6}. The experiment is indeed equivalent to the comparison between our model and the recently proposed model by \citet{Bertsimas2019Accounting} without multicollinearity constraints. We evaluate the models based on two criteria: (i) the solution time and (ii) the number of branch-and-bound nodes searched until the optimal solution is obtained. If the model with \eqref{f:4_6} is faster and searches less number of nodes, then we can show that \eqref{f:4_6} is effective. In this experiment, we use the cases where our proposed model is able to find the optimal solution within a given time limit in the previous experiments.

\noindent\textit{\textbf{Experiment 4: Alternative solution procedure vs. simple penalty function}}
We demonstrate the effectiveness of the alternative solution procedure by comparing it with a simple penalty function. The benchmark is obtained by replacing the alternative solution procedure with the simple penalty function $f_q$. The benchmark model will be referred to as $\mathcal{MP}_\text{penalty}(k)$.

\noindent\textit{\textbf{Experiment 5: Sensitivity analysis}}
In this experiment, we demonstrate how the solutions obtained from $\mathcal{MP}_{\text{lazy}}$ differ and lead to different validation measures when the penalty parameters $\lambda_E, \lambda_\pi, \lambda_{h},$ and $\lambda_{l}$ change. The Bodyfat dataset is used. We set $k=7,8,9,$ and $10$, excluding  $k=3,4,5,$ and $6$ as our model $\mathcal{MP}_{\text{lazy}}$ found feasible solutions (see the results of Experiment 1 in Section \ref{S:5.4}) and the penalty parameters did not affect the results in these cases because they are used to search for alternative solutions when feasible solutions are not available.  

To check the effect of $\lambda_E$, we run the algorithm for $\lambda_E \in \{0.5, 1, 1.5,...,4.5, 5\}$ with the other parameters fixed using the values presented in Section \ref{S:5.2}. Similarly, to check the effect of $\lambda_\pi$, we run the algorithm for $\lambda_\pi \in  \lbrace 0.025, 0.05,...,0.225, 0.25 \rbrace$ with the other parameters fixed. Because $E \in [0,1]$ and $\pi \in \lbrace 1,2,...,k \rbrace$ are on different scales, different values are checked for $\lambda_E$ and $\lambda_\pi$ in this experiment.
Note that, of the penalty parameters, we only vary $\lambda_\pi$ and $\lambda_E$. This is because the solutions from the experiment for Bodyfat do not exhibit heteroscedasticity (see the results for Bodyfat in Figure \ref{fig:asp}). Because the $p$-values are either very large or small, changing the associated penalty parameters $\lambda_h$ and $\lambda_l$ does not change $r_h$ and $r_l$ in this experiment.

\subsection{Experimental results} \label{S:5.4}

 We now present the experimental results. For the numerical experiments, we utilize Intel(R) Core(TM) i7-8700 CPU @ 3.40GHz (8 CPUs) and 32GB RAM. All models and algorithms are implemented with Python, in which mathematical models are solved using Gurobi 9.0.0. For the construction of the linear model in $\mathcal{FS}(k)$, we employ the Python package \emph{statsmodels} \cite{seabold2010statsmodels}. In every experiment, none of the solutions from the comparative models demonstrate linearity between their residuals and fitted values, so the corresponding results are not provided.   

\noindent\textit{\textbf{Results for Experiment 1: Proposed model vs. simple benchmark models}} To measure the explanatory power of each model, we introduce a measure for relative explanatory power $REP = \frac{R^2_{adj}(S)}{R^2_{adj}(S_{\mathcal{MP}_\text{base}})}$, where $S_{\mathcal{MP}_\text{base}}$ is a subset obtained by solving $\mathcal{MP}_\text{base}$ and $S$ is the solution of the corresponding model. We use the $R^2_{adj}$ of $\mathcal{MP}_\text{base}$ as a denominator of $REP$ since $\mathcal{MP}_\text{base}$ provides the greatest $R^2_{adj}$ of the compared models (except for LASSO) because it does not have any diagnostic constraints.  Note that $REP$ of LASSO can be greater than 1, because LASSO does not include the log transform constraint \eqref{f:3_4}.

 We present the summarized results for the datasets and the number of variables selected ($k$) in Tables \ref{table:result1} and \ref{table:result1_byp}, respectively. The complete results are available in the online supplement. In both tables, $REP$ is the average over the dataset or the $k$ values. To test the performance when using only feasible cases, we also present $REP_\text{feas}$, which only considers cases with feasible solutions available when calculating the average. Additionally, we count the number of cases satisfying (i) the $t$-test and residual test, (ii) the $t$-test only, and (iii) the residual test only.

\begin{table}[h]
  \centering
  \caption{Results by dataset}\adjustbox{max width=\textwidth}{
    \begin{tabular}{|c|c|c|c|c|c|c|c|c|c|c|c|c|}
    \hline
    \multirow{2}[4]{*}{Dataset} & \multicolumn{4}{c|}{$REP$}      & \multicolumn{4}{c|}{$REP_{feas}$} & \multicolumn{4}{c|}{Execution time} \\
\hline          & $\mathcal{MP}_{\text{lazy}}$ & $\mathcal{MP}_{\text{base}}$ & $\quad\mathcal{FS}\quad$ & LASSO & $\mathcal{MP}_{\text{lazy}}$ & $\mathcal{MP}_{\text{base}}$ & $\quad\mathcal{FS}\quad$ & LASSO & $\mathcal{MP}_{\text{lazy}}$ & $\mathcal{MP}_{\text{base}}$ & $\quad\mathcal{FS}\quad$ & LASSO \\
    \hline
    Housing & 0.950 & 1.000 & 1.000 & 0.979 & 0.950 & 1.000 & 1.000 & 0.979 & 329.676 & 7.802 & 0.066 & 0.001 \\
    Servo & 0.921 & 1.000 & 1.000 & 0.946 & 0.367 & 1.000 & 1.000 & 0.915 & 494.756 & 60.021 & 0.094 & 0.000 \\
    AutoMPG & 0.929 & 1.000 & 0.998 & 0.927 & 0.929 & 1.000 & 0.998 & 0.927 & 478.818 & 332.581 & 0.140 & 0.001 \\
    Automobiles & 0.992 & 1.000 & 1.000 & 0.943 & 0.992 & 1.000 & 1.000 & 0.943 & 549.824 & 513.232 & 0.348 & 0.001 \\
    Winequality & 0.872 & 1.000 & 1.000 & 0.997 & 0.658 & 1.000 & 1.000 & 1.000 & 487.790 & 41.402 & 0.058 & 0.008 \\
    Bodyfat & 0.783 & 1.000 & 1.000 & 1.002 & 0.816 & 1.000 & 1.000 & 1.002 & 469.123 & 22.906 & 0.062 & 0.004 \\
    Barro & 0.910 & 1.000 & 0.989 & 0.833 & 1.000 & 1.000 & 0.974 & 0.593 & 376.991 & 3.597 & 0.056 & 0.001 \\
    Carseats & 0.983 & 1.000 & 1.000 & 0.948 & 1.000 & 1.000 & 1.000 & 0.929 & 155.391 & 4.808 & 0.071 & 0.001 \\
    Crime & 0.847 & 1.000 & 0.993 & 0.887 & 0.823 & 1.000 & 0.992 & 0.872 & 420.495 & 23.823 & 0.141 & 0.003 \\
    Framing & 0.999 & 1.000 & 1.000 & 0.997 & 1.000 & 1.000 & 1.000 & 1.000 & 303.066 & 7.046 & 0.069 & 0.001 \\
    Griliches & 0.987 & 1.000 & 0.997 & 0.900 & 0.987 & 1.000 & 0.997 & 0.900 & 163.162 & 184.169 & 0.154 & 0.001 \\
    Hprice3 & 0.950 & 1.000 & 0.988 & 0.961 & 0.921 & 1.000 & 0.989 & 0.941 & 335.505 & 3.551 & 0.047 & 0.001 \\
    \hline
    \multirow{2}[4]{*}{Dataset} & \multicolumn{4}{c|}{$t$-test \& residual test} & \multicolumn{4}{c|}{$t$-test}   & \multicolumn{4}{c|}{Residual test} \\
\hline         & $\mathcal{MP}_{\text{lazy}}$ & $\mathcal{MP}_{\text{base}}$ & $\mathcal{FS}$ & LASSO & $\mathcal{MP}_{\text{lazy}}$ & $\mathcal{MP}_{\text{base}}$ & $\mathcal{FS}$ & LASSO & $\mathcal{MP}_{\text{lazy}}$ & $\mathcal{MP}_{\text{base}}$ & $\mathcal{FS}$ & LASSO \\
    \hline
    Housing & \textbf{8} & 1     & 1     & 1     & 8     & 8     & 8     & 5     & 8     & 1     & 1     & 1 \\
    Servo & 1     & 0     & 0     & 0     & 8     & 7     & 7     & 8     & 1     & 0     & 0     & 0 \\
    AutoMPG & \textbf{8} & 0     & 0     & 0     & 8     & 8     & 8     & 3     & 8     & 0     & 0     & 0 \\
    Automobiles & \textbf{8} & 3     & 5     & 0     & 8     & 8     & 8     & 1     & 8     & 3     & 5     & 0 \\
    Winequality & 3     & 0     & 0     & 0     & 7     & 5     & 5     & 4     & 3     & 0     & 0     & 0 \\
    Bodyfat & 4     & 0     & 0     & 0     & 4     & 0     & 0     & 1     & 8     & 0     & 0     & 0 \\
    Barro & 3     & 3     & 3     & 0     & 3     & 3     & 3     & 0     & 8     & 8     & 8     & 8 \\
    Carseats & 6     & 5     & 5     & 7     & 6     & 5     & 5     & 7     & 8     & 8     & 8     & 8 \\
    Crime & 7     & 0     & 0     & 1     & 8     & 8     & 8     & 1     & 7     & 0     & 0     & 2 \\
    Framing & 4     & 4     & 4     & 5     & 4     & 4     & 4     & 5     & 8     & 8     & 8     & 8 \\
    Griliches & \textbf{8} & 4     & 2     & 3     & 8     & 6     & 5     & 3     & 8     & 4     & 2     & 3 \\
    Hprice3 & 5     & 0     & 0     & 0     & 7     & 6     & 4     & 4     & 5     & 0     & 0     & 0 \\
    \hline
    \end{tabular}}%
  \label{table:result1}%
\end{table}%

\begin{table}[!h]
  \centering
  \caption{Results by $k$}\adjustbox{max width=\textwidth}{
    \begin{tabular}{|c|c|c|c|c|c|c|c|c|c|c|c|c|}
    \hline
    \multirow{2}[4]{*}{$k$} & \multicolumn{4}{c|}{$REP$}      & \multicolumn{4}{c|}{$REP_\text{feas}$} & \multicolumn{4}{c|}{Execution time} \\
\hline         & $\mathcal{MP}_{\text{lazy}}$ & $\mathcal{MP}_{\text{base}}$ & $\quad\mathcal{FS}\quad$ & LASSO & $\mathcal{MP}_{\text{lazy}}$ & $\mathcal{MP}_{\text{base}}$ & $\quad\mathcal{FS}\quad$ & LASSO & $\mathcal{MP}_{\text{lazy}}$ & $\mathcal{MP}_{\text{base}}$ & $\quad\mathcal{FS}\quad$ & LASSO \\
    \hline
    3     & 0.952 & 1.000 & 0.996 & 0.902 & 0.947 & 1.000 & 0.996 & 0.899 & 30.671 & 14.843 & 0.050 & 0.001 \\
    4     & 0.869 & 1.000 & 0.997 & 0.913 & 0.869 & 1.000 & 0.997 & 0.913 & 114.242 & 34.682 & 0.065 & 0.001 \\
    5     & 0.917 & 1.000 & 0.995 & 0.926 & 0.909 & 1.000 & 0.995 & 0.922 & 294.359 & 64.270 & 0.083 & 0.002 \\
    6     & 0.913 & 1.000 & 0.999 & 0.959 & 0.916 & 1.000 & 0.999 & 0.960 & 415.163 & 95.284 & 0.102 & 0.002 \\
    7     & 0.924 & 1.000 & 0.998 & 0.956 & 0.937 & 1.000 & 0.997 & 0.945 & 517.561 & 128.043 & 0.118 & 0.002 \\
    8     & 0.932 & 1.000 & 0.998 & 0.972 & 0.939 & 1.000 & 0.998 & 0.962 & 522.185 & 131.130 & 0.136 & 0.002 \\
    9     & 0.921 & 1.000 & 0.998 & 0.978 & 0.915 & 1.000 & 0.999 & 0.964 & 578.240 & 141.099 & 0.149 & 0.003 \\
    10    & 0.948 & 1.000 & 0.998 & 0.977 & 0.958 & 1.000 & 0.998 & 0.954 & 603.944 & 142.587 & 0.166 & 0.003 \\
    \hline
    \multirow{2}[4]{*}{$k$} & \multicolumn{4}{c|}{$t$-test \& residual test} & \multicolumn{4}{c|}{$t$-test}   & \multicolumn{4}{c|}{Residual test} \\
\hline{}          & $\mathcal{MP}_{\text{lazy}}$ & $\mathcal{MP}_{\text{base}}$ & $\mathcal{FS}$ & LASSO & $\mathcal{MP}_{\text{lazy}}$ & $\mathcal{MP}_{\text{base}}$ & $\mathcal{FS}$ & LASSO & $\mathcal{MP}_{\text{lazy}}$ & $\mathcal{MP}_{\text{base}}$ & $\mathcal{FS}$ & LASSO \\
    \hline
    3     & 11    & 4     & 4     & 5     & 12    & 11    & 11    & 9     & 11    & 4     & 4     & 6 \\
    4     & \textbf{12} & 4     & 4     & 3     & 12    & 11    & 11    & 7     & 12    & 4     & 4     & 5 \\
    5     & 11    & 4     & 4     & 3     & 12    & 11    & 11    & 7     & 11    & 4     & 4     & 4 \\
    6     & 9     & 3     & 3     & 2     & 11    & 10    & 10    & 6     & 10    & 4     & 4     & 3 \\
    7     & 7     & 3     & 2     & 2     & 9     & 9     & 8     & 5     & 10    & 5     & 4     & 3 \\
    8     & 6     & 1     & 1     & 1     & 9     & 7     & 5     & 3     & 9     & 4     & 4     & 3 \\
    9     & 5     & 0     & 1     & 1     & 8     & 5     & 5     & 3     & 9     & 3     & 4     & 3 \\
    10    & 4     & 1     & 1     & 0     & 6     & 4     & 4     & 2     & 8     & 4     & 4     & 3 \\
    \hline
    \end{tabular}}%
  \label{table:result1_byp}%
\end{table}%

In Table \ref{table:result1}, we observe that $\mathcal{MP}_\text{lazy}$ obtains solutions with a $REP$ of $90\%$ or above except for the Winequality, Bodyfat, and Crime datasets. This indicates that our model maintains explanatory power while satisfying all diagnostics constraints. Column $REP_\text{feas}$ also indicates that our model maintains an $R^2_{adj}$ closer to the base model. The results in the `$t$-test \& residual test' column indicate that our model is able to find substantially more linear models satisfying both the statistical significance of all coefficients and the residual assumptions for most of the datasets. In particular, our model finds optimal solutions satisfying all diagnostics constraints for all cases for four data sets (bolded in the `$t$-test \& residual test' column). On the other hand, our model finds the same number of feasible solutions as the benchmarks for the Framing dataset, with the results from our model for $REP_\text{feas}$ equal to 1 for this dataset. In fact, the corresponding solutions for $\mathcal{MP}_\text{base}$ are exactly the same as those of our model. This indicates that, although $\mathcal{MP}_\text{base}$ finds an optimal solution without considering any diagnostics or statistical significance, the solution fortunately has no insignificant coefficients and satisfies the residual assumptions. Thus, these results in Table \ref{table:result1} make clear that our model is able to provide a quality linear model independent of the dataset. Finally, the execution time indicates that our model can find linear models within a practical timeframe.  

Table \ref{table:result1_byp} presents the results by averaging or summing the number of variables selected ($k$). We can verify that the results of our model for $REP_\text{feas}$ are close to 1, and the results in the `$t$-test \& residual test' column are substantially better than those of the benchmarks, as in Table \ref{table:result1}. This suggests that our model can also generate quality subsets regardless of the size of $k$.  

\begin{figure}[h!]
	\centering
	\includegraphics[page=1,width=\textwidth]{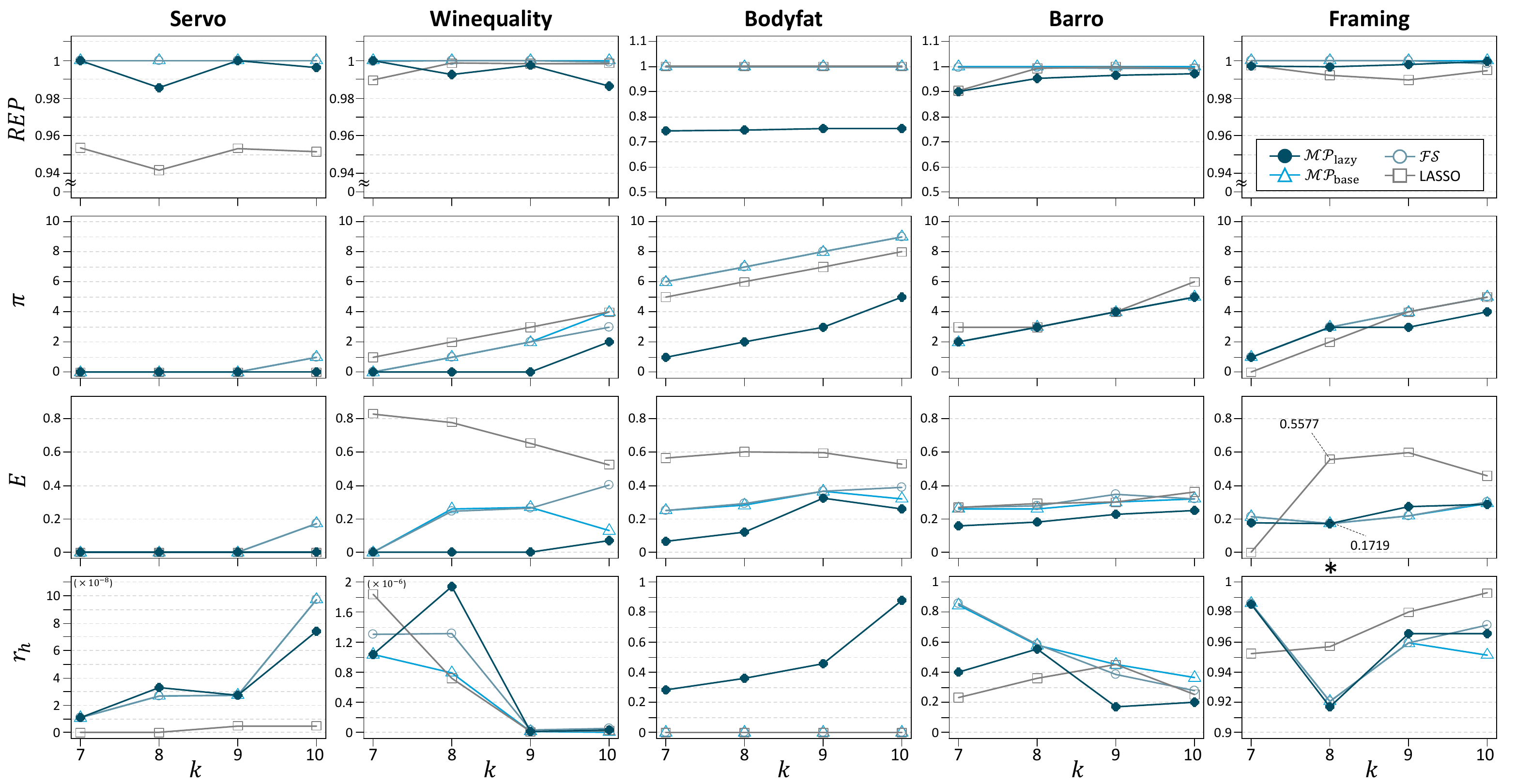}
	\caption{Comparison of the alternative and benchmark solutions. In each plot, the horizontal axis is the number of selected variables ($k$).}\label{fig:asp}
\end{figure}

 In Figure \ref{fig:asp}, we check the results for the alternative solution procedure for the following cases where no feasible solution is found within the 600-second time limit: Servo, Winequality, Bodyfat, Barro, and Framing with $k = 7,8,9,10$. In the plot matrix of Figure \ref{fig:asp}, the horizontal and vertical axes represent the datasets and the performance measures, respectively. Note that $E=0$ if a solution has no statistically insignificant coefficient because, as illustrated in Section \ref{S:4}, a penalty term is activated when a violation of the corresponding statistical test occurs. Also, heteroscedasticity would not be a concern for the Barro and Framing dataset because every case in the dataset satisfies the regression assumption ($r_h > 0.01$).

The plots in Figure \ref{fig:asp} explicitly illustrate the promising performance of the alternative solution procedure. As seen in the charts in the first row, the $REP$ of the proposed model remains above 0.89 except for the Bodyfat dataset. Moreover, except for the two cases (Framing with $k=7,8$) including the case marked with a `$*$', the measures $\pi$, $E$, and $r_h$ for the alternative solutions are better than or equal to those of the benchmarks. These results indicate that the alternative solutions improve upon those of the benchmarks in most cases. For example, for Winequality ($k=8$), the $\pi$ and $E$ of $\mathcal{MP}_\text{lazy}$ are considerably lower than those of the benchmarks while it has $REP$ greater than 0.99. On the other hand, the $REP$ for the alternative solutions for Bodyfat is relatively less than for the other results. However, $\pi$ and $E$ are significantly better than the benchmarks. This indicates that our model sacrifices explanatory power (i.e., MSE) to improve the statistical tests and diagnostics ($\pi$ and $E$).

We investigate the case marked with a `$*$' (Framing with $k = 8$) more in detail, in which the alternative solutions do not outperform the benchmark solutions for every measure. This is due to our algorithm's tolerance parameter $\tau$. Let $S'_{\text{current}}$ be the solution of $\mathcal{MP}_\text{lazy}$ with $\pi = 3$ and $E = 0.1719$ and $S'_\text{new}$ be the solution for LASSO with $\pi = 2$ and $E = 0.5577$  from the result in Figure \ref{fig:asp}. These two solutions can be compared using our alternative solution procedure. Based on Algorithms \ref{alg:asp} and \ref{alg:dt}, $f_t$ is called in Step 4 of Algorithm \ref{alg:asp}. Then Step 1 of Algorithm \ref{alg:dt} compares  $E(S'_\text{new})$ and $E(S'_{\text{current}})$ accounting for the given tolerance parameter $\tau=0.1$. Because $E(S'_\text{new}) - E(S'_{\text{current}}) = 0.3858 > 0.1$, the alternative solution procedure concludes that $S'_{\text{current}}$ is a better solution, which is a clearly reasonable decision. We find that LASSO gives a better solution in only one case, Framing with $k = 7$.

\begin{figure}[!h]
	\centering
	\includegraphics[page=1,width=\textwidth]{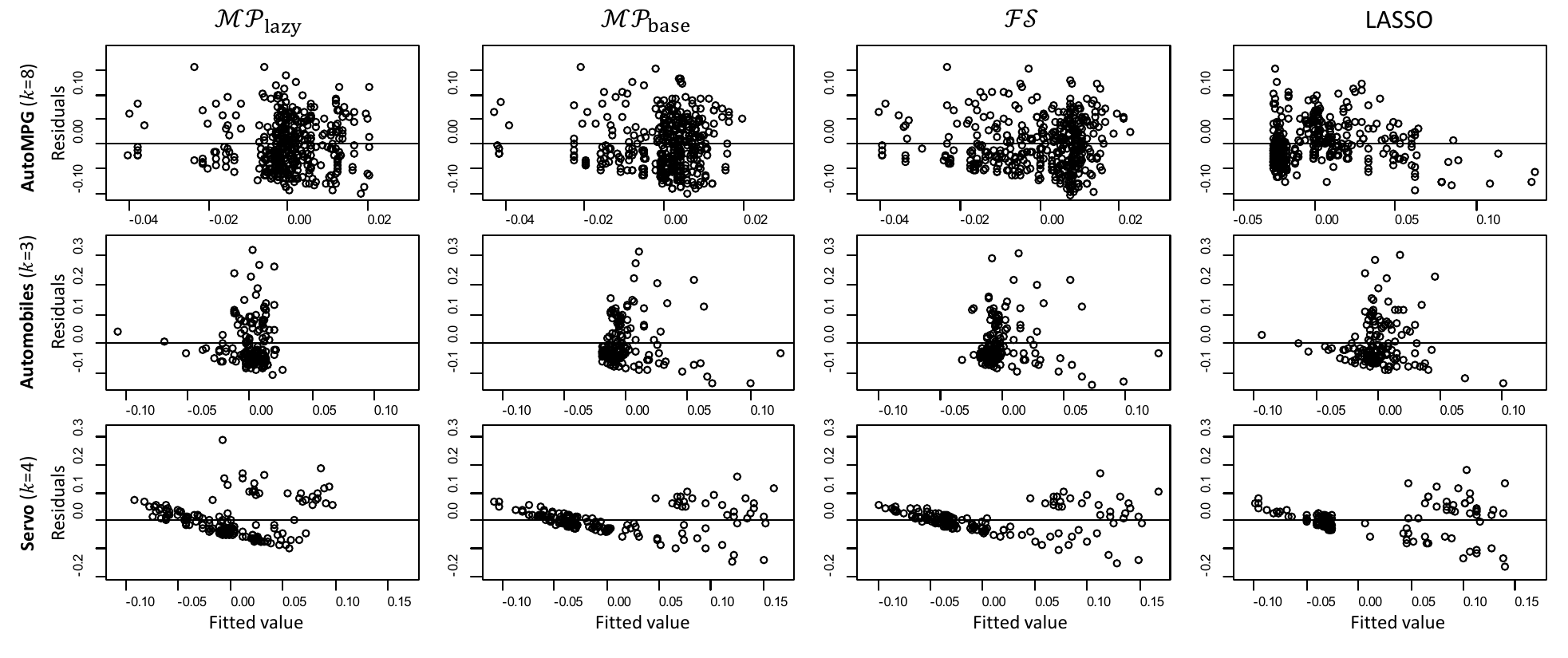}
	\vspace{-0.4cm}
	\caption{Residual plots for three representative cases}\label{fig:result_residual}
\end{figure}

Figure \ref{fig:result_residual} presents representative residual plots (residuals versus fitted values) for the AutoMPG ($k=8$), Automobile ($k=3$), and Servo ($k=4$).  In Table \ref{table:heteropvalue}, the corresponding $p$-values are presented. The plots in Figure \ref{fig:result_residual} show that the variance of the residuals from the benchmarks gradually increases or decreases, while the linear regression model derived from our model has a relatively consistent residual trend in relation to the fitted values. The $p$-values in Table \ref{table:heteropvalue} show that our model provides better solutions because a higher $p$-value is preferable for the heteroscedasticity tests.

\begin{table}[!h]
\caption{$p$-values from heteroscedasticity tests}
	\begin{center}
		\noindent\adjustbox{max width=0.8\textwidth}{%
\begin{tabular}{|c|c|c|c|c|}
				\hline
				\multicolumn{1}{|c|}{Case}  	&  \multicolumn{4}{c|}{$r_h$} 	\\
				\cline{2-5}
				\multicolumn{1}{|c|}{(dataset, $k$)} & $\quad\>\mathcal{MP}_\text{lazy}\quad\>$	&  $\mathcal{MP}_\text{base}$	& $\mathcal{FS}$ & LASSO \\
				\hline
				AutoMPG ($k=8$)		& 0.0142		& $2.6402\times10^{-5}$	&  $2.7821\times10^{-8}$ &	$2.8149\times 10^{-12}$  \\	
				Automobile ($k=3$)	& 0.0111		& $3.3471\times10^{-6}$	& $1.1835\times10^{-9}$ &	$1.6423\times10^{-6}$\\
				Servo ($k=4$)		& 0.0106		& $1.4686 \times 10^{-11}$	& $1.5836\times10^{-16}$ & $7.1300 \times 10^{-24}$	\\
				\hline
		\end{tabular}}\label{table:heteropvalue}
	\end{center}
\end{table}

\noindent\textit{\textbf{Results for Experiment 2: Proposed model vs. iterative model}} In Table \ref{table:result2_iter}, we compare our model with the iterative model ($\mathcal{MP}_\text{iter}$) for cases where the solution for $\mathcal{MP}_\text{base}$ violates some of the tests and diagnostics. We only consider these 28 cases because both algorithms are not needed if $\mathcal{MP}_\text{base}$ can provide a solution that satisfies all tests and diagnostics. In the Status column, `alt sol' indicates that, within the time limit, a model could not find a solution with statistically significant coefficients for all of the selected variables, and an alternative solution is obtained at termination. The terms `opt' and `best feas' represent cases where an optimal solution (satisfying all constraints) and a feasible solution are found, respectively. The term `infeas' for $\mathcal{MP}_\text{lazy}$ means that the problem is infeasible and our model returns an alternative solution. These cases are organized according to the following hierarchy: `opt' $>$ `best feas' $>$ `alt sol' $=$ `infeas'. In columns Status, $E$, and $\pi$, the results indicating that our model is better are in boldface. The results in Table \ref{table:result2_iter} are summarized in Table \ref{table:result2_summary}.

\begin{table}[htbp]
  \centering
  \caption{Comparative results for the iterative model}
    \noindent\adjustbox{max width=1\textwidth}{
    \begin{tabular}{|c|c|c|c|c|c|c|c|c|c|c|c|}
    \hline
    \multirow{1}[4]{*}{dataset} & \multirow{1}[4]{*}{$k$} & \multicolumn{2}{c|}{Status} & \multicolumn{2}{c|}{$REP$} & \multicolumn{2}{c|}{$E$} & \multicolumn{2}{c|}{$\pi$} & \multicolumn{2}{c|}{Execution time} \\
\cline{3-12}          &       & $\mathcal{MP}_\text{iter}$ &$\mathcal{MP}_\text{lazy}$ & $\mathcal{MP}_\text{iter}$ &$\mathcal{MP}_\text{lazy}$ & $\mathcal{MP}_\text{iter}$ &$\mathcal{MP}_\text{lazy}$ & $\mathcal{MP}_\text{iter}$ &$\mathcal{MP}_\text{lazy}$ & $\mathcal{MP}_\text{iter}$ &$\mathcal{MP}_\text{lazy}$ \\
    \hline
    Servo & 10    & alt sol & \textbf{best feas} & 1     & 0.9963 & 0.1745 & \textbf{0} & 1     & \textbf{0} & 3600  & 3601 \\
    \hline
    Bodyfat & 3     & opt & opt   & 0.9998 & 0.9998 & 0     & 0     & 0     & 0     & 83    & \textbf{5} \\
    Bodyfat & 4     & alt sol & \textbf{opt}   & 0.9997 & 0.7557 & 0.2541 & \textbf{0} & 1     & \textbf{0} & 3600  & \textbf{86} \\
    Bodyfat & 5     & alt sol & \textbf{opt}   & 0.9999 & 0.7539 & 0.162 & \textbf{0} & 4     & \textbf{0} & 3600  & \textbf{1978} \\
    Bodyfat & 6     & alt sol & \textbf{best feas} & 0.9999 & 0.7568 & 0.3183 & \textbf{0} & 5     & \textbf{0} & 3600  & 3601 \\
    Bodyfat & 7     & alt sol & \textbf{best feas} & 0.9999 & 0.7414 & 0.343 & \textbf{0} & 6     & \textbf{0} & 3600  & 3601 \\
    Bodyfat & 8     & alt sol & alt sol & 1     & 0.7524 & 0.2904 & \textbf{0.1619} & 7     & \textbf{1} & 3600  & 3601 \\
    Bodyfat & 9     & alt sol & alt sol & 1     & 0.7587 & 0.2898 & \textbf{0.1403} & 8     & \textbf{4} & 3600  & 3601 \\
    Bodyfat & 10    & alt sol & alt sol & 1     & 0.76  & 0.3554 & \textbf{0.3026} & 9     & \textbf{3} & 3600  & 3601 \\
    \hline
    Barro & 6     & alt sol & alt sol & 0.4900  & 0.5191 & 0.5495 & \textbf{0.0931} & 3     & \textbf{1} & 3600  & 3600 \\
    Barro & 7     & alt sol & alt sol & 0.9664 & 0.5372 & 0.3176 & \textbf{0.1095} & 3     & \textbf{2} & 3600  & 3601 \\
    Barro & 8     & alt sol & alt sol & 0.9829 & 0.5171 & 0.4093 & \textbf{0.1817} & 3     & 3     & 3600  & 3600 \\
    Barro & 9     & alt sol & alt sol & 0.9905 & 0.9469 & 0.4361 & \textbf{0.2407} & 4     & 4     & 3600  & 3600 \\
    Barro & 10    & alt sol & alt sol & 0.9892 & 0.9721 & 0.391 & \textbf{0.2507} & 5     & 5     & 3600  & 3601 \\
    \hline
    Winequality & 8     & alt sol & \textbf{opt}   & 0.9959 & 0.9924 & 0.4756 & \textbf{0} & 1     & \textbf{0} & 3600  & \textbf{88} \\
    Winequality & 9     & alt sol & \textbf{opt}   & 0.9988 & 0.9974 & 0.5346 & \textbf{0} & 2     & \textbf{0} & 3600  & \textbf{48} \\
    Winequality & 10    & alt sol & alt sol & 0.9988 & 0.9863 & 0.4835 & \textbf{0.0725} & 3     & \textbf{2} & 3600  & 3600 \\
    \hline
    Carseats & 8     & opt   & opt   & 0.9998 & 0.9998 & 0     & 0     & 0     & 0     & 397   & \textbf{9} \\
    Carseats & 9     & alt sol & \textbf{opt}   & 0.8117 & 0.9923 & 0.5056 & \textbf{0} & 3     & \textbf{0} & 3600  & \textbf{289} \\
    Carseats & 10    & alt sol & alt sol & 1     & 0.9502 & 0.2099 & \textbf{0.1743} & 2     & \textbf{1} & 3600  & 3601 \\
    \hline
    Framing & 7     & alt sol & alt sol & 0.9886 & 0.9971 & 0.2128 & \textbf{0.1768} & 2     & \textbf{1} & 3600  & 3601 \\
    Framing & 8     & alt sol & alt sol & 0.9643 & 0.8653 & 0.2711 & \textbf{0.1649} & 4     & \textbf{3} & 3600  & 3602 \\
    Framing & 9     & alt sol & alt sol & 0.9965 & 0.8651 & 0.2213 & \textbf{0.1614} & 4     & 4     & 3600  & 3601 \\
    Framing & 10    & alt sol & alt sol & 0.9969 & 0.9654 & 0.3761 & \textbf{0.1957} & 5     & 5     & 3600  & 3601 \\
    \hline
    Griliches & 9     & alt sol & \textbf{opt}   & 1     & 0.9963 & 0.0509 & \textbf{0} & 1     & \textbf{0} & 3600  & \textbf{417} \\
    Griliches & 10    & alt sol & \textbf{opt}   & 0.9996 & 0.9934 & 0.0604 & \textbf{0} & 2     & \textbf{0} & 3600  & \textbf{843} \\
    \hline
    Hprice3 & 9     & opt   & opt   & 0.9952 & 0.9952 & 0     & 0     & 0     & 0     & 625   & \textbf{47} \\
    Hprice3 & 10    & alt sol & alt sol & 0.9914 & 0.9936 & 0.553 & \textbf{0.0943} & 3     & \textbf{1} & 3600  & 3600 \\
    \hline
    \multicolumn{2}{|c|}{Average} &    -   &-       & 0.9698 & 0.8699 & 0.2945 & \textbf{0.0900} & 3.25  & \textbf{1.43} & 3253.7 & \textbf{2450.9} \\
    \hline
    \end{tabular}}%
  \label{table:result2_iter}%
\end{table}%

 \begin{table}[h!]
 \caption{Summary table for the results in Table \ref{table:result2_iter}}
	\centering
	\noindent\adjustbox{max width=0.7\textwidth}{%
		\begin{tabular}{|>{\centering}p{3cm}|S[table-format=3.1, table-column-width=0.15\textwidth]| S[table-format=3.1, table-column-width=0.15\textwidth]| S[table-format=3.1, table-column-width=0.15\textwidth]|}
			\hline\rule{0pt}{2.5ex}
			Measures & {$\mathcal{MP}_\text{iter}$ wins} 	&{$\mathcal{MP}_\text{lazy}$ wins}	& {Tie}	\\
			\hline 		\rule{0pt}{2.5ex}			
			Status 			&0				& 10				& 18	\\
			$E$				&0				& 25			& 3	\\
			$\pi$			&0				& 20			& 8	\\
			Execution time 	&0				& 10				& 18	\\
			\hline
	\end{tabular}}
	\label{table:result2_summary}
\end{table}

According to Tables \ref{table:result2_iter} and \ref{table:result2_summary}, $\mathcal{MP}_\text{lazy}$ outperforms $\mathcal{MP}_\text{iter}$ in most cases, while maintaining an explanatory power around 87\% of $\mathcal{MP}_\text{base}$. In cases where $\mathcal{MP}_\text{lazy}$ finds an optimal solution, $\mathcal{MP}_\text{lazy}$ is appreciably faster than $\mathcal{MP}_\text{iter}$. Moreover, by comparing performance measures $E$ and $\pi$ for the alternative solutions, we can see that the alternative solutions from our model outperform those from the iterative model in terms of the statistical significance of the coefficients, while differences in the explanatory power can be ignored in most cases. Furthermore, for Barro ($k=6$), Carseats ($k=9$), Framing ($k=7$), and Hprice3 ($k=10$), our model finds a better solution in terms of both explanatory power and statistical significance. Several results in the Bodyfat and Barro datasets indicate that the explanatory power of our model is less than 80\% of $\mathcal{MP}_\text{base}$. However, $\mathcal{MP}_\text{lazy}$ clearly outperforms $\mathcal{MP}_\text{iter}$ in terms of the statistical significance of the coefficients in these cases.

\noindent\textit{\textbf{Results for Experiment 3: Proposed model vs. model without the relaxed $t$-test constraints}} To confirm the effectiveness of the proposed relaxed $t$-test constraint \eqref{f:4_6}, we compare our proposed model with and without \eqref{f:4_6}. We run the models on the cases (dataset and $k$ pairs) where the optimal solution is found within 600 seconds. For each case, we run the algorithms 30 times to obtain the average and median number of explored nodes and computation time. Table \ref{table:summary_ex3} provides summary statistics by aggregating the results by $k$, where the complete results are reported in the  online supplement.

\begin{table}[h!]
\centering
\caption{Summary table for the results of experiment 3}
\noindent\adjustbox{max width=0.6\textwidth}{
\begin{tabular}{|c|c|c|c|c|c|}
\hline
\multicolumn{2}{|c|}{} & \multicolumn{2}{c|}{\#node} & \multicolumn{2}{c|}{Time} \\ \hline
$k$        & \#case      & Ratio      & \%improved      & Ratio     & \%improved    \\ \hline
3        & 12          & \textbf{0.998}      & 72.7           & 1.051     & 18.2         \\ 
4        & 12          & \textbf{0.984}      & 63.6           & 1.015     & 27.3         \\ 
5        & 5           & \textbf{0.862}      & 80.0           & \textbf{0.940}     & 60.0         \\ 
6        & 5           & \textbf{0.821}      & 100           & \textbf{0.892}     & 100         \\ 
7        & 2           & \textbf{0.918}      & 100           & \textbf{0.972}     & 50         \\ 
8        & 2           & \textbf{0.831}      & 100           & \textbf{0.893}     & 50         \\ \hline
ALL      & 39          & \textbf{0.936}      & 77.8           & \textbf{0.989}    & 41.7         \\ \hline
$k\geq 5$      & 14          & \textbf{0.846}      & 92.3           & \textbf{0.921}    & 71.4         \\ \hline
\end{tabular}}\label{table:summary_ex3}
\end{table}

For each case (dataset and $k$ pair), we define Ratio to be the average ratio between the model with and without the cut. If Ratio is less than 1, it implies that the model with the cut is better for the corresponding performance measure (either \#node and Time). In Table \ref{table:summary_ex3}, for each $k$, Ratio is the average Ratio out of all corresponding cases and \%improved is the percentage of cases with Ratio $< 1$. The last two rows include the averages across all cases and cases with $k \geq 5$.

In Table \ref{table:summary_ex3}, we can verify the advantage of including the relaxed $t$-test constraint. For every $k$, the Ratio for \#node is less than 1. This shows that our model searches less number of solutions before obtaining the optimal solution than the model without the constraint. This is because the constraint reduces the solution space by cutting some solutions with insignificant coefficients. As a result, the computation time to obtain the optimal solution decreases as shown in the Time column. In particular, the computation time of the model with the cut is significantly faster when $k$ is larger (e.g., $k = 5,...,8$): by including the constraint, \#node and computation time are significantly decreased by 15.4\% and 7.9\%, respectively. This trend implies that for a larger $k$, more solutions are excluded by the constraint. It corresponds to the fact that it is more likely to have insignificant coefficients when we have more explanatory variables.   

\noindent \textit{\textbf{Results for Experiment 4: Alternative solution procedure vs. simple penalty function}}
In Table \ref{table:onlypenalty1}, the $p$-values for the coefficients from the linear models derived from $\mathcal{MP}_\text{lazy}$ and $\mathcal{MP}_\text{penalty}$ for Bodyfat ($k=7$) and Carseat ($k=10$) are presented, where the two cases are selected because the two approaches provide clearly distinguished solutions. Recall that the only difference between  $\mathcal{MP}_\text{lazy}$ and $\mathcal{MP}_\text{penalty}$ is that the former uses the alternative solution procedure whereas the latter uses the simple penalty function $f_q$.  The $p$-values greater than $1-\alpha_{E} = 1 - 0.95 = 0.05$ are bolded. 

\begin{table}[h!] 
\caption{$p$-values from the $t$-tests for the coefficients}

	\begin{center}
		\noindent\adjustbox{max width=\textwidth}{%
			\begin{tabular}{|c|c|c|c|c c c c c c c c c c|}
				\hline
				Dataset ($k$) & Model	& $R^2_{adj}$ 		& $E$ 	&\multicolumn{10}{c|}{$p$-values}  \\
				\hline 
				\multirow{2}{*}{Bodyfat (7)} &\multicolumn{1}{l|}{$\mathcal{MP}_\text{lazy}$}	&0.7288	&0.0649	&0.0000	&\textbf{0.0649}	&0.0112	&0.0001	&0.0076	&0.0045	&0.0091 & & &			\rule{0pt}{2.5ex}	\\
				&\multicolumn{1}{l|}{$\mathcal{MP}_\text{penalty}$}	& 0.7274	&0.0899	&\textbf{0.0899}	&0.0000	&0.0035	&0.0402	&0.0302	&0.0050	&0.0179 & & &	\\
				\hline 
				\multirow{2}{*}{Carseat (10)} &\multicolumn{1}{l|}{$\mathcal{MP}_\text{lazy}$}	&0.8296	&0.1736	&0.0000	&0.0000	&0.0002	&0.0000	&0.0000	&0.0000	&\textbf{0.1736}	&0.0000	&0.0000	&0.0000
			\rule{0pt}{2.5ex}	\\
				  &\multicolumn{1}{l|}{$\mathcal{MP}_\text{penalty}$}	& 0.8682	&0.1800	&0.0000	&0.0000	&0.0000	&0.0000	&0.0004	&\textbf{0.1800}	&0.0000	&0.0000	&0.0000	&\textbf{0.1800}\\
				\hline
		\end{tabular}}\label{table:onlypenalty1}
	\end{center}
\end{table}
Note that the average of the violating $p$-values ($E$) for $\mathcal{MP}_\text{lazy}$ for Bodyfat ($k=7$) is 0.0649, while it is 0.0899 for $\mathcal{MP}_\text{penalty}$. Also, the $R^2_{adj}$ of $\mathcal{MP}_\text{lazy}$ is greater than that of $\mathcal{MP}_\text{penalty}$. In Carseat ($k=10$), the average of the violating $p$-values for $\mathcal{MP}_\text{lazy}$ is smaller than that of $\mathcal{MP}_\text{penalty}$ ($0.1736 < 0.1800 $) and furthermore the number of insignificant coefficients for $\mathcal{MP}_\text{lazy}$ is smaller than  that of $\mathcal{MP}_\text{penalty}$ ($1<2$). The $R_{adj}^2$ of $\mathcal{MP}_\text{lazy}$ is worse, however, the difference in $R^2_{adj}$ is negligible. Thus, our model, which includes the alternative solution procedure, can provide a better solution than the model employing a simple penalty function.

\begin{figure}[h]
\centering
		\begin{subfigure}[b]{0.45\textwidth}
		\centering	\includegraphics[page=1,width=\textwidth]{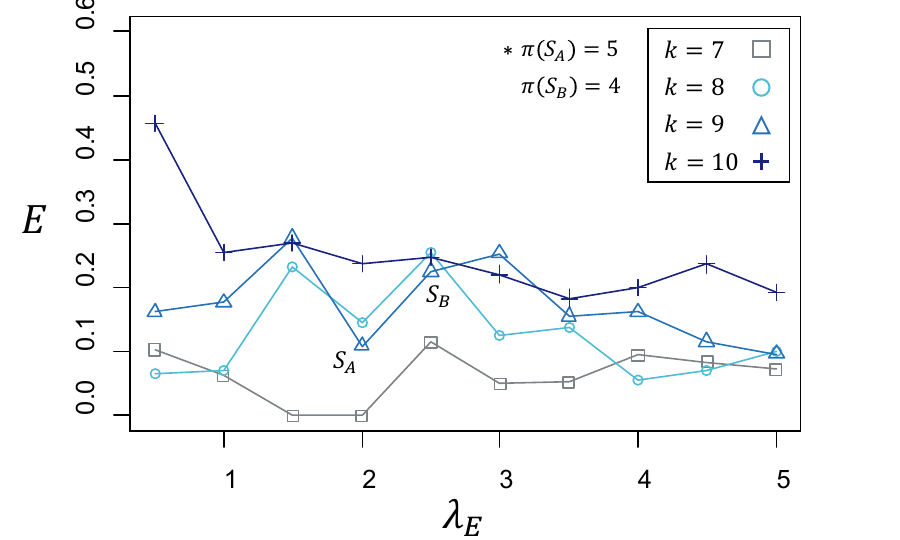} 
			\caption{$E$ (Average violating $p$-value)} \label{fg_exp_effect1}
		\end{subfigure}
		\begin{subfigure}[b]{0.45\textwidth}
		\centering
	\includegraphics[page=2, width=\textwidth]{sensitivity_analysis.pdf} 
			\caption{$\pi$ (Number of insignificant coefficients)} \label{fg_exp_effect2}
		\end{subfigure}
	\caption{Trends in the measures according to changes in the parameters (Bodyfat with $k=7,8,9,10$)}
	\label{fig6.1}
\end{figure}

\noindent\textbf{\textit{Results for Experiment 5: Sensitivity analysis}}
The plots in Figure \ref{fig6.1} represent the trends in $E$ and $\pi$ according to changes in $\lambda_E$ and $\lambda_\pi$ for the Bodyfat dataset, respectively. In each plot, it can be observed that each violation measure decreases with an increase in the corresponding parameter. This relationship does not appear to be monotonous, because the optimization algorithm embedded in the solver searches different spaces and terminates due to the time limit every time it is called. For example, let us compare the solutions $S_A$ and $S_B$ in Figure \ref{fg_exp_effect1}. (The other validation measures are not significantly different.). According to each value of $E$ and $\pi$, it is obvious that $S_A$ is a better solution than $S_B$ under the parameter set with $\lambda_E = 2$ and $\lambda_E = 2.5$. However, because the optimization solver did not find $S_A$ during the search with $\lambda_E = 2.5$, $S_B$ is returned as the best alternative solution. In Figure \ref{fg_exp_effect2}, it is noteworthy that the decrease in $\pi$ does not appear when $\lambda_\pi$ is greater than 0.15. This indicates that increasing the parameter does not necessarily improve the validation measure after a certain point. 

\section{Conclusion}
\label{S:6}
 Although minimizing fitting errors is the most important objective when building a regression model, it is also important to check if all the  regression assumptions and diagnostics are met or satisfied. However, most of the current approaches do not (or cannot) incorporate this model validation step into their algorithms. In this work, we propose a fully automated model building procedure for multiple linear regression subset selection that integrates model building and validation based on mathematical programming. The procedure is further improved in terms of computation efficiency by including the relaxed $t$-test constraint which can be efficiently derived.  The proposed procedure is also capable of providing an alternative solution when there is no feasible solution or a feasible solution is not found within the set time limit. 

The computational experiments with real-world datasets show that our model can provide quality solutions satisfying all of the considered diagnostics while maintaining an $R^2_{adj}$ value close to the base solution (optimal value when all the diagnostics are ignored). Furthermore, the results demonstrate the viability of our model in real regression analysis by setting a practical time limit. We also find that the alternative solutions are superior to those of the benchmarks. The $R^2_{adj}$ value of the alternative solutions are comparable with those of the benchmarks, while being significantly better in terms of diagnostics. We also show that our model based on lazy constraints outperforms the existing iterative approach. The proposed lazy constraint-based procedure is faster than the iterative approach while providing more quality alternative solutions within the time limit. Lastly, we demonstrate the benefit of the relaxed $t$-test constraint showing that including the constraint can significantly decrease the number of explored solutions and the computation time.

In conclusion, we strongly believe that our model is useful in practice because we integrate model building (subset selection) and validation steps, whereas traditional approaches require human decisions and significant trial-and-error to alternate the two steps. Furthermore, the proposed lazy constraint approach based on mathematical programming that addresses the statistical tests and diagnostics for linear regression is substantially more efficient than the existing iterative model. We also remark that other regression diagnostics that are not considered in this paper can be easily integrated into our lazy constraint framework. One of the most challenging research directions for the improvement of the current procedure includes formulating a linear or a convex constraint to replace the current approximation constraint for $t$-tests or improve the quality of the approximated constraint.

\section*{Acknowledgements}

This research was partly supported by the National Research Foundation of Korea (NRF) grant funded by the Korea Government (Ministry of Science, ICT \& Future Planning) (No. NRF-2015R1C1A1A02036682) and was also supported in part by the Korea Institute for Advancement of Technology (KIAT) grant funded by the Korea Government (MOTIE) (The Competency Development Program for Industry Specialist) under Grant P0008691.  

\bibliographystyle{plainnat}
\bibliography{ref}

\end{document}